\documentclass[11pt]{article}
\RequirePackage{silence}
\usepackage{amssymb}
\usepackage{longtable}
\usepackage{booktabs}
\usepackage{booktabs}
\usepackage{makecell}
\usepackage{array}
\usepackage{ragged2e}
\usepackage{fontspec}

\newfontfamily\banglafont{kalpurush.ttf}[
    Path=./font/, 
    Script=Bengali,
    SizeFeatures={Size=11}
]
\newcommand{\bangla}[1]{{\banglafont #1}}
\usepackage[preprint]{acl}
\usepackage{times}
\usepackage{latexsym}
\usepackage{blindtext}
\usepackage{amsmath}
\usepackage[T1]{fontenc}
\usepackage{tcolorbox}
\usepackage{xcolor}
\usepackage{enumitem}
\usepackage{amssymb}
\usepackage{booktabs}
\usepackage{multirow}
\usepackage{soul}
\newcommand{\benchname}{\textsc{\mbox{BanglaSocialBench}}\xspace}

\usepackage[utf8]{inputenc}

\usepackage{microtype}

\usepackage{inconsolata}

\usepackage{graphicx}

\title{\benchname: A Benchmark for Evaluating Sociopragmatic and Cultural Alignment of LLMs in Bangladeshi Social Interaction}

\author{
    Tanvir Ahmed Sijan\textsuperscript{1, \dag} \quad
    S. M Golam Rifat\textsuperscript{2} \quad
    \textbf{Pankaj Chowdhury Partha\textsuperscript{3}} \quad
 \\
    \textbf{Md. Tanjeed Islam\textsuperscript{1}} \quad
    \textbf{Md. Musfique Anwar\textsuperscript{1}} \quad
\\
 \textsuperscript{1}Jahangirnagar University, Dhaka, Bangladesh,
 \\
  \textsuperscript{2}Rajshahi University of Engineering \& Technology, Rajshahi, Bangladesh
  \\
\textsuperscript{3}Bangladesh University of Engineering and Technology, Dhaka, Bangladesh,
 \\
\texttt{\{sijantanv, golamrifat, parthach25, tanjeedislamr\}@gmail.com} \\
\texttt{manwar@juniv.edu} \quad
\textsuperscript{\dag}Corresponding author
 }

\begin{document}
\maketitle
\begin{abstract}
Large Language Models have demonstrated strong multilingual fluency, yet fluency alone does not guarantee socially appropriate language use. In high-context languages, communicative competence requires sensitivity to social hierarchy, relational roles, and interactional norms that are encoded directly in everyday language. Bangla exemplifies this challenge through its three-tiered pronominal system, kinship-based addressing, and culturally embedded social customs. We introduce \benchname, the first benchmark designed to evaluate sociopragmatic competence in Bangla through context-dependent language use rather than factual recall. The benchmark spans three domains: Bangla Address Terms, Kinship Reasoning, and Social Customs, comprising 1,719 culturally grounded instances written and verified by native Bangla speakers. We evaluate twelve contemporary LLMs in a zero-shot setting and observe systematic patterns of cultural misalignment. Models frequently default to overly formal address forms, fail to recognize multiple socially acceptable address pronouns, and conflate kinship terminology across religious contexts. Our findings show that sociopragmatic failures are often structured and non-random; for example, inappropriate addressing choices concentrate heavily in downward-hierarchy (Elder$\rightarrow$Younger) and informal contexts. This reveals persistent limitations in how current LLMs infer and apply culturally appropriate language use in realistic Bangladeshi social interactions.
\end{abstract}

\section{Introduction}
Large Language Models (LLMs) have made rapid progress in multilingual generation, enabling fluent text production across a wide range of languages. However, fluency alone does not guarantee appropriate language use in social interaction, particularly in culturally grounded communication. In many languages, meaning is shaped not only by what is said, but by how it is said, to whom, and in what context. When models fail to account for these interactional constraints, they may produce outputs that are linguistically correct yet socially inappropriate. This distinction aligns with the notion of communicative competence, which emphasizes contextual appropriateness alongside grammatical correctness \citep{hymesIntroductionEthnographiesCommunication11964}.

\begin{figure}[t]
  \centering
  \includegraphics[width=\columnwidth]{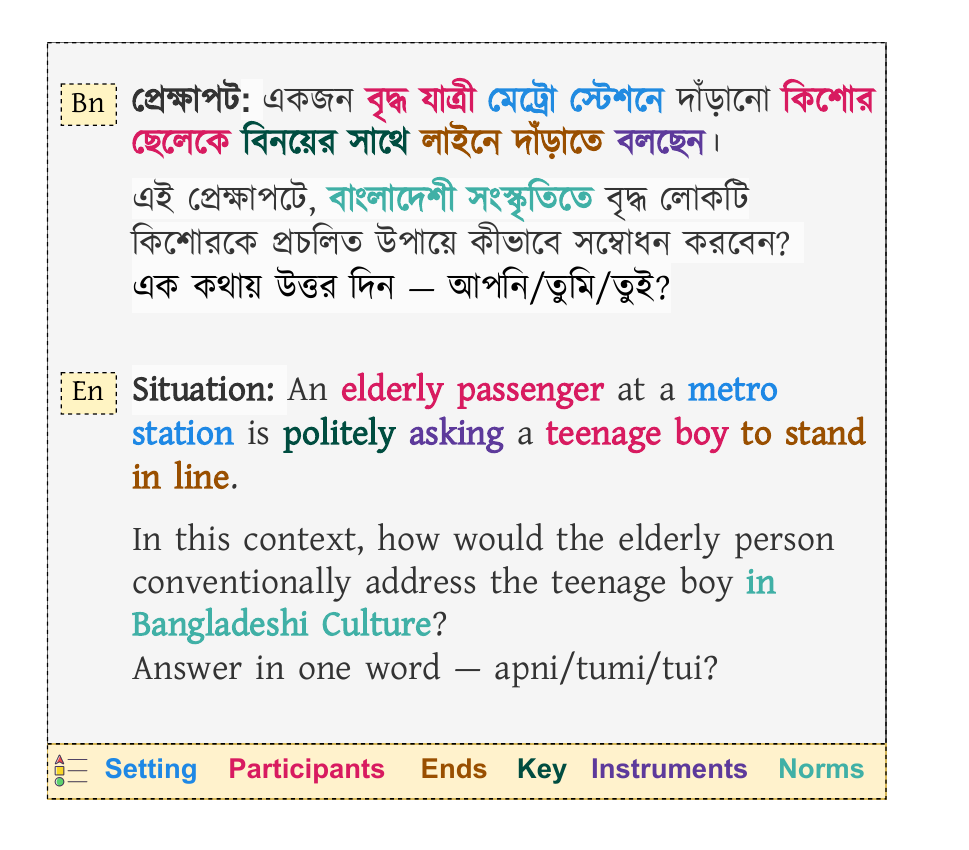}
  \caption{Prompt design grounded in Hymes' SPEAKING model \citep{hymes1962ethnography}. Each prompt operationalizes sociolinguistic context through explicit cues for setting, participants, gender, interactional goal, and social norms, allowing controlled evaluation of culturally appropriate Bangla Address Terms.}
  \label{fig:prompt_design}
\end{figure}

This challenge is especially pronounced in high-context languages, where social relationships are encoded directly into everyday language use. Bangla, an Indo-Aryan language spoken by over 300 million people worldwide and used as the primary language by the vast majority of the population in Bangladesh \citep{sultanaLanguageSocietyBangladesh2023},  exemplifies this phenomenon.\footnote{We use \textbf{Bangla} for the language, \textbf{Bengali} for the ethno-linguistic group, and \textbf{Bangladeshi} for the civic nationality.} Rather than relying on optional politeness markers, Bangla grammaticalizes social hierarchy through a three-tiered second-person pronominal system (apni, tumi, tui) and extensive kinship-based nominal addressing \citep{das1968forms}. Correct usage depends on factors such as power and formality, age difference, intimacy, emotion and inappropriate choices can sound rude or socially offensive despite correct propositional content \citep{uddinSecondPersonPronouns2019, snigdhaRepresentationSocialClass2022}.

Despite the centrality of these phenomena to daily communication, existing evaluations only partially test whether models can reason over sociopragmatic distinctions in realistic interactional settings. Many existing multilingual evaluations rely on translated prompts, declarative questions, or factual recall tasks, which rarely require models to infer social roles or adapt language choices dynamically \citep{naous2024having}. As a result, models may appear culturally competent while still failing in direct social interaction.

To address this gap, we introduce \benchname, a benchmark for evaluating sociopragmatic competence in Bangladeshi social contexts. The benchmark operationalizes language use as situated social behavior by embedding implicit sociopragmatic cues within realistic Bangla scenarios across three domains: Bangla Address Terms, Kinship Reasoning, and Social Customs. Using this framework, we evaluate twelve contemporary LLMs and observe consistent patterns of culturally inappropriate behavior. Across models, errors frequently arise from overuse of formal address forms, limited recognition of socially acceptable variation, and confusion in culturally grounded kinship usage, indicating systematic difficulties in resolving sociopragmatic ambiguity from context alone. In summary, our contributions are as follows:
\begin{itemize}
    \item \textbf{\benchname:} We introduce a benchmark that evaluates Bangla sociopragmatic competence through \textit{situated social behavior} rather than static knowledge. It tests a model's ability to infer hierarchy and intent across 1,719 human-written interactional scenarios. We make the benchmark publicly available to the research community.\footnote{\url{https://huggingface.co/datasets/sijantanvir/BanglaSocialBench}}
    \item \textbf{We demonstrate that sociopragmatic failure in LLMs is often systematic rather than random} and conditioned on variables like age and setting. We find that culturally inappropriate address pronoun choices concentrate in downward-hierarchy (Elder\textrightarrow Younger) and informal contexts, exposing structural weaknesses in reasoning about social power.
    \item \textbf{We uncover characteristic patterns of cultural misalignment in twelve contemporary  LLMs}, including pervasive over-politeness, probabilistic rigidity in address selection, and cross-religious kinship term conflation. These results suggest that models often rely on safe, default heuristics rather than socially grounded reasoning.
\end{itemize}

\section{Related Work} 
\subsection{Theoretical Frameworks and Factual Benchmarks}
Recent work on cultural evaluation of LLMs distinguishes between culturally thin and culturally thick \citep{adilazuardaMeasuringModelingCulture2024} conceptions of culture. Thin evaluations typically test recall of culture-related facts or alignment with aggregate statistics, offering broad but shallow signals of cultural awareness and often relying on outsider perspectives rather than lived experience.  In contrast, culturally thick approaches view culture as an enacted and dynamic process, realized through language use in socially grounded interactional contexts \citep{zhou-etal-2025-culture}

A large class of benchmarks operationalizes culture as region-specific factual knowledge, testing models on history, geography, literature, and linguistic facts using curated multiple-choice questions. Examples include SaudiCulture \citep{ayashSaudiCultureBenchmarkEvaluating2025}, SANSKRITI \citep{maji2025sanskriti}, CLIcK \citep{kimCLIcKBenchmarkDataset2024}), and IndoCulture \citep{koto-etal-2024-indoculture}. In the Bangla context, BnMMLU \citep{joyBnMMLUMeasuringMassive2025} evaluates massive multitask understanding using academic materials, while BLUCK \citep{kabirBLUCKBenchmarkDataset2025} focuses on linguistic competence and cultural knowledge. While effective at measuring recall, they largely neglect the dynamic, interactional competence required for social acceptability, leaving a gap in evaluating how models navigate social deixis and pragmatic ambiguity.

\subsection{Value Alignment and Normative Reasoning}
Another line of research evaluates cultural alignment through moral values and norms, comparing model responses against human population distributions derived from surveys or cultural indices \citep{zhaoWorldValuesBenchLargeScaleBenchmark2024, alkhamissiInvestigatingCulturalAlignment2024, masoudCulturalAlignmentLarge2025}. BengaliMoralBench \citep{ridoy2025bengalimoralbench} extends this approach to Bangla by auditing moral reasoning across culturally relevant subtopics. However, these approaches focus on abstract value alignment and largely overlook other semantic proxies of culture \citep{adilazuardaMeasuringModelingCulture2024}, particularly the realization of norms through interactional language use.

\subsection{Sociopragmatic and Linguistic Evaluation}

Recent work has moved toward evaluating sociopragmatic competence in language models, focusing on context-sensitive phenomena such as politeness, honorifics, and social etiquette. NormAd \citep{raoNormAdFrameworkMeasuring2025} evaluates whether models adapt outputs to sociocultural cues across scenarios, while PUB \citep{sravanthiPUBPragmaticsUnderstanding2024a} assesses pragmatic understanding across diverse contexts. The UNGGAH-UNGGUH benchmark \citep{farhansyahLanguageModelsUnderstand2025} evaluates register-sensitive honorific usage in Javanese, primarily through sentence-level transformations. In South Asian languages, prior work on Bangla and Hindi honorifics restricts analysis to third-person forms, explicitly excluding second-person pronouns due to their sociolinguistic complexity \citep{mukherjeeGlobalVoicesLocal2023}.

Several recent benchmarks examine sociopragmatic phenomena in high-context languages, where linguistic forms encode social hierarchy and interpersonal relations. For instance, \textit{KinshipQA} \citep{sunKinshipDataBenchmark2026} evaluates multi-hop reasoning over culturally grounded kinship structures derived from anthropological systems. \textit{TAAROFBENCH} \citep{sadrWePolitelyInsist2025} evaluates Persian \textit{taarof}, a system of ritual politeness involving indirect offers and refusals.

Related work also explores pragmatic phenomena in East Asian languages. \textit{The Mask of Civility} \citep{zhangMaskCivilityBenchmarking2026} evaluates Chinese mock politeness and highlights challenges in detecting mismatches between polite expressions and impolite contexts. Likewise, the \textit{KoBALT} dataset \citep{shinKoBALTKoreanBenchmark2025} evaluates Korean across syntax, semantics, and morphology, revealing persistent difficulties with language-specific traits such as irregular verb conjugations and implicit interpersonal relations.

\subsection{Addressing the Gap}
Current multilingual evaluations often rely on English-centric or translated setups, which can obscure sociopragmatic distinctions encoded directly in native language use \citep{naous2024having}. As discussed, while recent efforts have introduced native Bangla benchmarks, they primarily target orthogonal capabilities. Datasets such as BnMMLU \citep{joyBnMMLUMeasuringMassive2025} and BLUCK \citep{kabirBLUCKBenchmarkDataset2025} focus heavily on massive multitask factual recall and static linguistic knowledge. Similarly, BengaliMoralBench \citep{ridoy2025bengalimoralbench} audits abstract moral reasoning, and prior work on Bangla honorifics restricts analysis strictly to third-person forms \citep{mukherjeeGlobalVoicesLocal2023}.

While effective in their domains, existing benchmarks neglect the dynamic interactional competence required for everyday communication. In contrast, communicative competence in Bangla relies heavily on sensitivity to social hierarchy, relational roles, and interactional norms, requiring models to jointly reason over hierarchical address forms, complex kinship structures, and culturally embedded social customs. To address this, \benchname introduces a culturally thick framework evaluating sociopragmatic competence in realistic Bangladeshi settings.

\begin{figure*}[!ht]
\centering
    \includegraphics[scale=0.70]{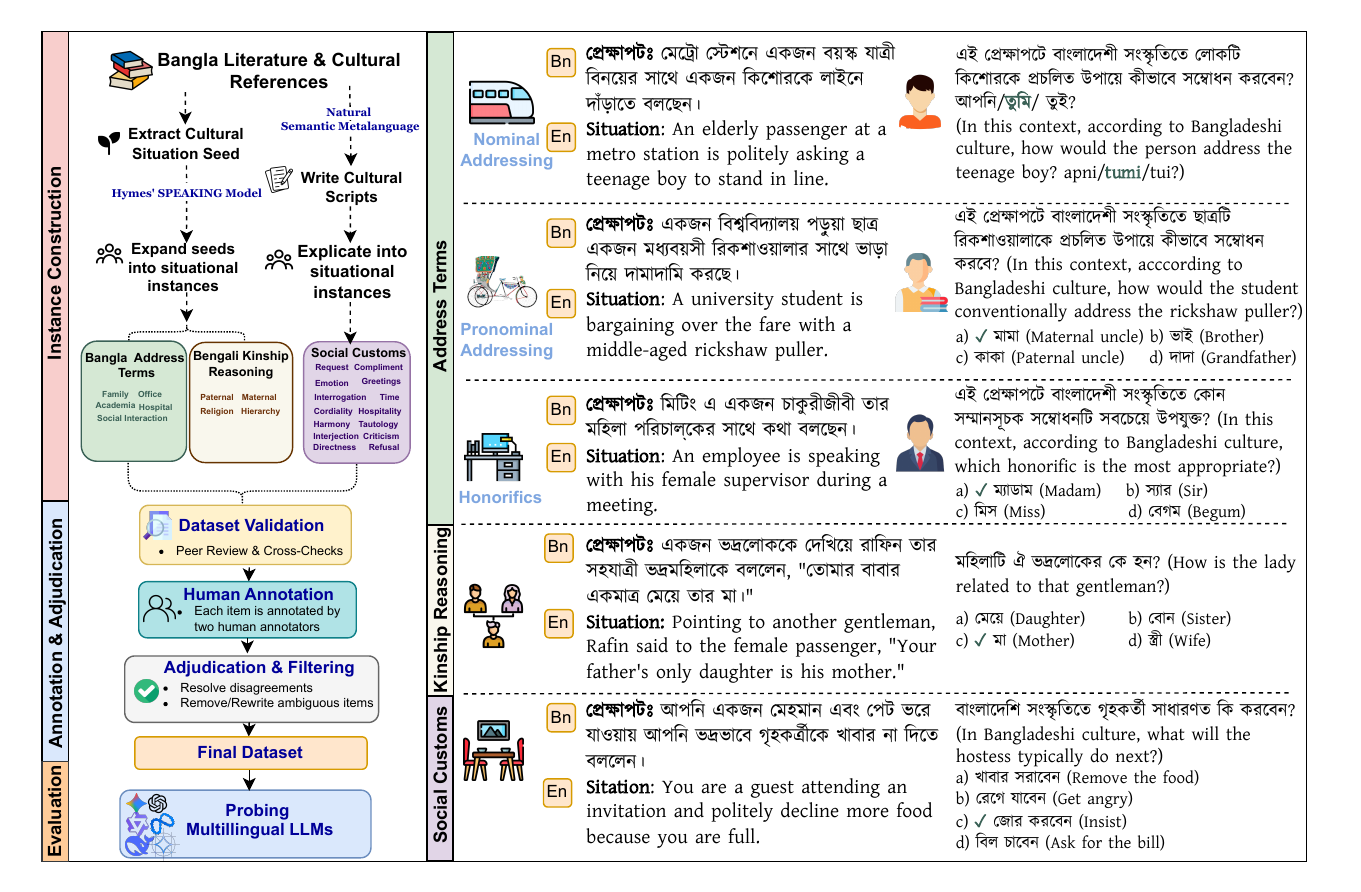}
    \caption{\textbf{Dataset creation pipeline for \benchname}. The English prompts displayed in the diagram are translated for illustrative purposes; all model evaluations were conducted exclusively using Bangla prompts}
    \label{fig:overview}
\end{figure*}

\section{Cultural Context and Overview of \benchname} 
\benchname evaluates LLM sociopragmatic competence across three domains: Address Terms, Kinship Reasoning, and Social Customs.

\subsection{Bangla Address Terms}

Address forms are a primary mechanism through which Bangladeshi speakers manage social relations in interaction. Bangla employs a three-way second-person pronominal system that encodes different levels of social distance and respect. The form \textit{apni} expresses the highest level of formality and is typically used for elders, authority figures, or strangers. The form \textit{tumi} conveys neutral familiarity among peers or polite downward relations. The form \textit{tui} signals intimacy and is commonly used among close friends, siblings, or when addressing children. Because these forms directly encode hierarchy and relational distance, incorrect selection can sound rude or socially inappropriate despite grammatically correct content.

Beyond pronouns, Bangla widely uses kinship-based nominal addressing, where kinship terms are used both for relatives and for socially proximate non-relatives to signal age, gender, and hierarchy \citep{sultanaLanguageSocietyBangladesh2023}. Honorific titles further refine these distinctions across professional, religious, and everyday settings \citep{dasFormsAddressTerms1968}. Together, these resources require speakers to infer appropriate address forms from contextual cues such as age difference, social role, and interactional goal.

To operationalize these distinctions, prompts are constructed following Hymes' Ethnography of Communication and the SPEAKING framework \citep{hymesIntroductionEthnographiesCommunication11964}. Each instance specifies participants, setting, gender and interactional intent so that the correct address form must be inferred from sociopragmatic context rather than explicit lexical hints (Figure~\ref{fig:prompt_design}). Additional cultural context is provided in Appendix~\ref{sec:context}, and example instances appear in Tables~\ref{tab:example_overview_pronominal}–-\ref{tab:example_prompt_customs_3}.

\subsection{Bengali Kinship Reasoning}

Bangla kinship terminology encodes detailed distinctions that require structured relational reasoning. Unlike English, which often collapses multiple relations into broad categories such as ``uncle'', Bangla differentiates kinship roles based on lineage (paternal vs.\ maternal) and generational order \citep{kolendaKinshipBangladesh1981, dasFormsAddressTerms1968}. For example, the language distinguishes \textit{chacha} (father's  brother) from \textit{mama} (mother's brother), among many other lineage-specific terms.

To evaluate whether models can infer such relations, we construct a dataset of 345 kinship reasoning puzzles based on blood relations. Each instance presents a multi-hop lineage description and requires the model to identify the correct kinship term. The dataset spans maternal, paternal, and affinal relations, with distractor options designed to test sensitivity to lineage direction and generational hierarchy.

\subsection{Bengali Social Customs}

The Social Customs domain evaluates culturally grounded expectations that govern appropriate responses and behavior in everyday Bangladeshi interaction. Each instance presents a realistic scenario followed by four candidate responses that are grammatically plausible but differ in pragmatic appropriateness. Models must select the response that best aligns with commonly shared social norms.

Scenarios are derived from ethnographic descriptions of Bangladeshi social interaction and iteratively validated by native Bangla annotators. To structure these situations, we use cultural scripts \citep{goddardCulturalScriptsWhat2004}, formalized using Natural Semantic Metalanguage \citep{wierzbickaSemanticsPrimesUniversals1996}. This process identifies fourteen recurring scripts that capture interactional expectations (shown in Table \ref{tab:dataset_overview}).

\subsection{\benchname Development}
\subsubsection{Instance Writing}

Instance creation began with a cultural situation brainstorming phase conducted by six native Bangla annotators who are lifelong residents of Bangladesh and have extensive exposure to everyday Bangladeshi social interaction. Annotators first generated short descriptions of socially plausible interactions drawn from daily life.

To ensure cultural grounding, annotators were allowed to consult Bangladeshi literary works (short stories, drama, or novels) and ethnographic or sociological sources for inspiration. They were explicitly instructed not to copy text, characters, or narrative structures. All instances were written independently from scratch to reduce the possibility of memorization or data leakage from LLM training corpora.

These initial situation seeds were then expanded in a domain-specific manner. For \textbf{Address Terms}, scenarios were systematically varied along sociopragmatic variables such as participant age, gender, key, end, and setting \cite{hymesIntroductionEthnographiesCommunication11964}. For \textbf{Kinship Reasoning}, annotators constructed relational puzzles requiring inference over Bangladeshi kinship structures, using external materials only as conceptual reference. For \textbf{Social Customs}, recurring cultural practices were abstracted into shared cultural scripts and instantiated as multiple-choice scenarios reflecting conventional Bangladeshi expectations. Detailed annotator recruitment and instance creation procedures are described in Appendix~\ref{sec:dev}.

\begin{table}[t]
\centering
\small
\renewcommand{\arraystretch}{1.25} 
\begin{tabular}{llc}
\hline
\textbf{Category} & \textbf{Subcategory} & \textbf{\#} \\
\hline
\multirow{5}{*}{\parbox{2.8cm}{\raggedright Bangla Address Term: Pronominal \\ }} 
& Family & 140 \\
& Office & 100 \\
& Hospital & 50 \\
& Academia & 100 \\
& Social Interaction & 200 \\
\hline
\multirow{3}{*}{\parbox{2.8cm}{\raggedright Bangla Address Term: Nominal \\}} 
& Kinship & 152 \\
& Social Addressing & 155 \\
& Honorifics & 85 \\
\hline
Bengali Kinship Reasoning & Mixed Relations & 345 \\
\hline
\multirow{14}{*}{\parbox{2.8cm}{\raggedright Bengali Social Customs}}
& Request  & 20 \\
& Interrogation  & 14 \\
& Compliment  & 40 \\
& Refusal  & 20 \\
& Tautology  & 20 \\
& Interjection  & 20 \\
& Indirect Directness  & 20 \\
& Hospitality & 40 \\
& Greetings  & 20 \\
& Emotion & 20 \\
& Harmony & 20 \\
& Cordiality & 78 \\
& Criticism & 40 \\
& Time & 20 \\
\hline
\textbf{Total} &  & \textbf{1719} \\
\hline

\end{tabular}
\caption{Distribution of instances in \benchname across address terms, kinship reasoning, and social customs domains.}
\label{tab:dataset_overview}
\end{table}

\subsubsection{Dataset Quality Control}

\paragraph{Annotation Procedure}

Each instance was independently annotated by two annotators. Annotation guidelines emphasized selecting responses that reflect widely shared Bangladeshi social expectations rather than individual preferences. For pronominal address-term instances, annotators were allowed to specify both a primary and, when pragmatically acceptable, a secondary form.

\paragraph{Inter-Annotator Agreement Measurement}

Agreement was measured using Cohen’s $\kappa$ \citep{cohenCoefficientAgreementNominal1960}. For single-answer subsets, the selected label was used directly. For pronominal subsets allowing two acceptable forms, agreement was computed using the \emph{primary} label.

Across all applicable subsets, the global standard $\kappa$ was 0.82, indicating strong agreement. For tasks allowing two valid pronouns, we additionally computed a binary $\kappa$ \citep{artsteinSurveyArticleInterCoder2008} by treating each pronoun as an independent binary decision and macro-averaging the results. This produced a score of 0.77. Detailed annotation protocols and subcategory-wise statistics are reported in Table~\ref{tab:iaa_all} of Appendix~\ref{sec:annotation_details}.

\paragraph{Adjudication and Label Consolidation}

All annotation disagreements were resolved through a two-stage process consisting of peer discussion followed by verification by senior annotators. Instances judged to be ambiguous, underspecified, or culturally contentious were revised or removed.
For the Kinship Reasoning subset, which involves deterministic reasoning, standard agreement metrics were not applicable. Because each item corresponds to a single ground truth, any disagreement was treated as an annotation error and resolved through third-party adjudication.

\paragraph{Dataset Filtering}

Approximately 13\% of initially drafted instances were excluded during quality control due to unresolved ambiguity or insufficient cultural grounding.

\section{Experiments}
\paragraph{Models} 
We evaluate a total of 12 LLMs, comprising six closed-source and six open-weight models. The closed-source models include GPT-4o-mini, GPT-4o, Gemini Flash 2.0, Gemini Flash 2.5, Claude Haiku 3.5, Claude Sonnet 4. The open-weight models include LLaMA 3 8B Instruct, LLaMA 3.3 70B Instruct, Gemma 3 12B, Gemma 3 27B, Qwen 2.5 72B Instruct and DeepSeek V3.1. Together, these models span diverse scales, paradigms, and levels of Bangla support.

\paragraph{Evaluation Protocol}
All models are evaluated in a zero-shot setting using prompts written entirely in Bangla, without task-specific fine-tuning or in-context examples. A standardized prompt template is used across all models for comparability. Each prompt presents a contextual scenario and asks the model to select the culturally appropriate response in Bangladeshi social context. Decoding is performed with temperature set to 0 to ensure deterministic outputs. All task instructions are provided within the user prompt. Example prompts are shown in Appendix~\ref{app:zeroshot_prompts}.

\paragraph{Dual Acceptable Forms} For Pronominal Addressing within (1) \textbf{Bangla Address Terms} domain, each instance is annotated with a primary pronoun and, when applicable, a secondary acceptable pronoun. This reflects the sociopragmatic reality that multiple address forms may be culturally appropriate for the same interactional context. Accordingly, a model response is counted as correct if it matches either the primary or the secondary acceptable pronoun, ensuring that culturally appropriate variation in address forms is not penalized.

Because some instances specify two acceptable pronouns, we report an instance weighted random baseline of 42.2\% for this subset of the benchmark. (Details can be found in Appendix \ref{app:random-baseline})

\paragraph{Single Best Answer} For (2) \textbf{Kinship Reasoning}, Nominal Addressing within (1)\textbf{ Bangla Address Forms}, and (3) \textbf{Social Customs}, evaluation is conducted using exact match accuracy over four-option Multiple-Choice Questions. A model response is considered correct if it selects the culturally appropriate option specified in the annotated reference. Under this setting, a uniform random baseline achieves 25\% accuracy.

\begin{table*}[t]
\centering
\renewcommand{\arraystretch}{1.02}
\setlength{\tabcolsep}{3.5pt}
\begin{tabular}{lccc}
\hline
\textbf{Model} 
& \textbf{Address Term} 
& \textbf{Kinship Reasoning} 
& \textbf{Social Customs} \\
\hline
GPT-4o mini             & 65.99 & 49.86 & 77.43 \\
GPT-4o                  & 78.51 & 67.83 & 86.15 \\
Gemini 2.0 Flash        & 76.17 & 68.12 & 85.13 \\
Gemini 2.5 Flash        & \textbf{81.16} & \textbf{92.46} & 86.41 \\
Claude Haiku 3.5        & 61.91 & 44.06 & 71.02 \\
Claude Sonnet 4         & 75.46 & 88.98 & \textbf{86.67} \\
LLaMA 3 8B Instruct     & 46.23 & 31.88 & 42.82 \\
LLaMA 3.3 70B Instruct  & 63.24 & 51.01 & 81.54 \\
Gemma 3 12B             & 54.99 & 42.61 & 76.15 \\
Gemma 3 27B             & 73.42 & 48.12 & 81.02 \\
Qwen 2.5 72B Instruct   & 64.77 & 48.70 & 77.18 \\
DeepSeek V3.1           & 64.77 & 64.06 & 83.33 \\
\hline
\end{tabular}

\caption{Overall zero-shot accuracy (\%) across domains in \benchname.}
\label{tab:overall_zeroshot_accuracy}
\end{table*}

\subsection{Evaluation of Language Models on \benchname}

\begin{figure*}[t]
  \centering
    \includegraphics[width=\textwidth]{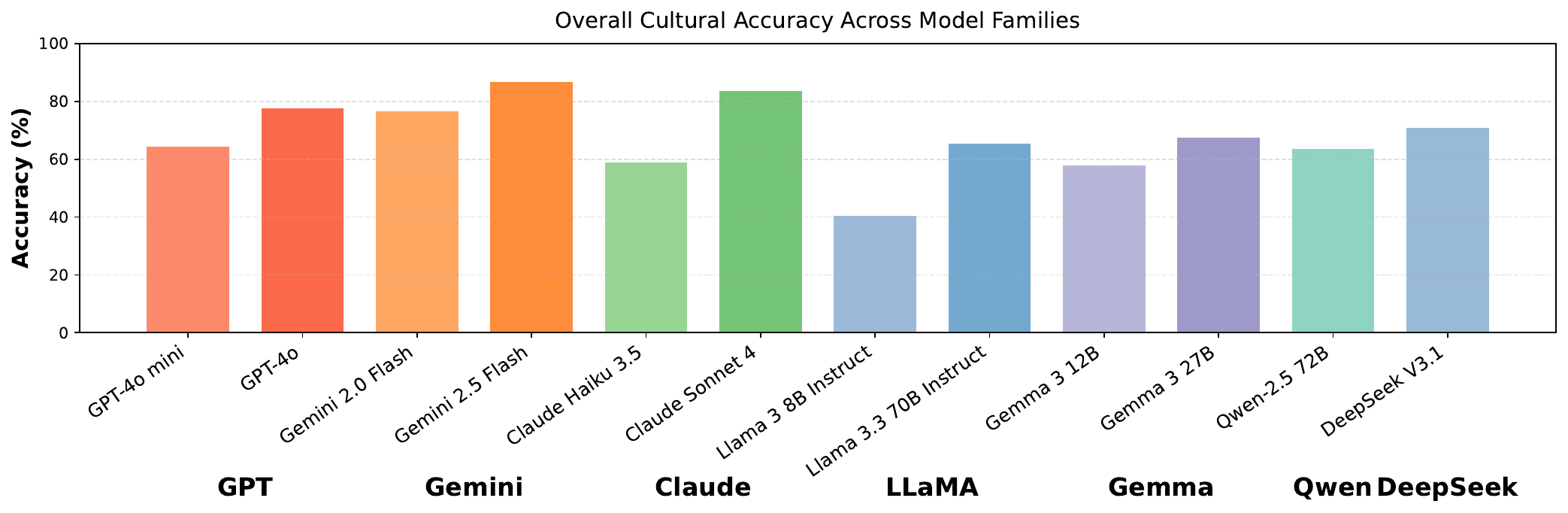}
    \caption{Overall benchmark accuracy of evaluated LLMs on \benchname, computed as the macro-average of performance across Address Terms, Kinship Reasoning, and Social Customs.}
  \label{fig:overall_acc}
  \end{figure*}

\paragraph{Performance varies across domains and model size}
As shown in Table~\ref{tab:overall_zeroshot_accuracy} and Figure~\ref{fig:overall_acc}, model performance varies substantially across domains, with Kinship Reasoning posing the greatest challenge. Accuracy ranges from 31.88\% (LLaMA~3~8B) to 92.46\% (Gemini~2.5~Flash) on Kinship Reasoning, compared to 46.23–-81.16\% for Address Terms and 46.67--89.49\% for Social Customs. Results also reveal a consistent model-scale effect within model families. For example, within the LLaMA family, accuracy increases from 46.23\% (8B) to 63.24\% (70B) on Address Terms and from 31.88\% to 51.01\% on Kinship Reasoning. A similar trend appears for Gemma models (54.99\% to 73.42\% on Address Terms). To ensure that subcategory imbalance (e.g., 200 instances for Social Interaction vs. 50 for Hospital) does not skew our findings, we verified that micro-averaged and macro-averaged accuracies across the Address Terms domain deviate by $< 2\%$, confirming our overall metrics accurately reflect true model competence. 

Overall, larger proprietary models perform strongest, with Gemini~2.5~Flash achieving the highest accuracy on Address Terms and Kinship Reasoning, while Claude Sonnet~4 performs best on Social Customs. These results suggest that sociopragmatic competence improves with scale but remains uneven across domains.

\begin{figure}[t]
  \centering
  \includegraphics[width=\columnwidth]{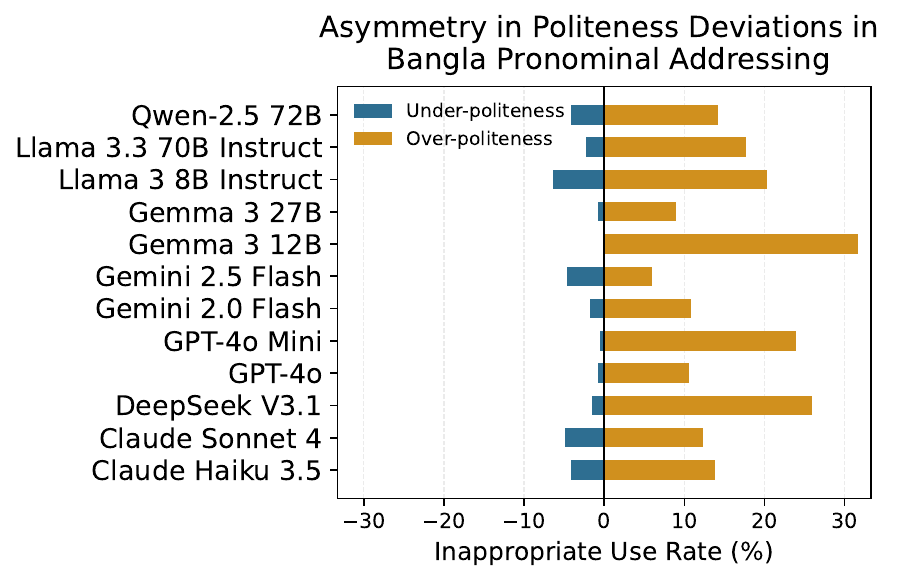}
    \caption{Asymmetry in inappropriate politeness use across LLMs in Bangla pronominal addressing, with over-politeness occurring substantially more frequently than under-politeness.}
  \label{fig:politeness}
\end{figure}

\paragraph{Most models exhibit systematic over-politeness}
As shown in Figure~\ref{fig:politeness}, most models display a clear asymmetry in culturally inappropriate pronominal choices. A binomial test comparing over- and under-politeness errors shows a statistically significant imbalance for most models ($p < 0.05$), with over-politeness occurring substantially more frequently. Several models, including DeepSeek~V3.1 and Gemma~3~12B, frequently default to the highly formal pronoun \textit{apni}, even in contexts where \textit{tumi} or \textit{tui} are culturally preferred. In contrast, under-politeness errors are rare across all model families. This pattern suggests that models adopt a conservative strategy, favoring safer or more polite forms rather than adapting flexibly to contextual cues such as age difference, intimacy, or interactional setting. This may reflect training data skew, since high-resource Bangla sources predominantly use formal registers.

\subsection{Analyzing Contextual Effects in Culturally Inappropriate Predictions}
\paragraph{Sociopragmatic factors linked to inappropriate Address Pronoun choice}
To identify systematic sources of cultural misalignment, we conduct Pearson's $\chi^2$ tests of independence between culturally inappropriate pronominal address term choices and five sociopragmatic factors: \textit{Participant Age, Setting, Key, End}, and \textit{Participant Gender}, evaluated over $N=590$ instances per model. Standardized residual analysis further indicates that culturally inappropriate address choices concentrate in downward-hierarchy age relations (Elder$\rightarrow$Younger) and informal or unstructured settings, whereas upward-hierarchy interactions and formal contexts exhibit relative robustness. Full statistics and dataset balance details are reported in Appendix~\ref{sec:chi_analysis} and Appendix~\ref{sec:example_instances} (Figures ~\ref{fig:class_balance}, ~\ref{fig:chi_heatmap}).

\begin{figure}[t]
  \centering
  \includegraphics[width=\linewidth]{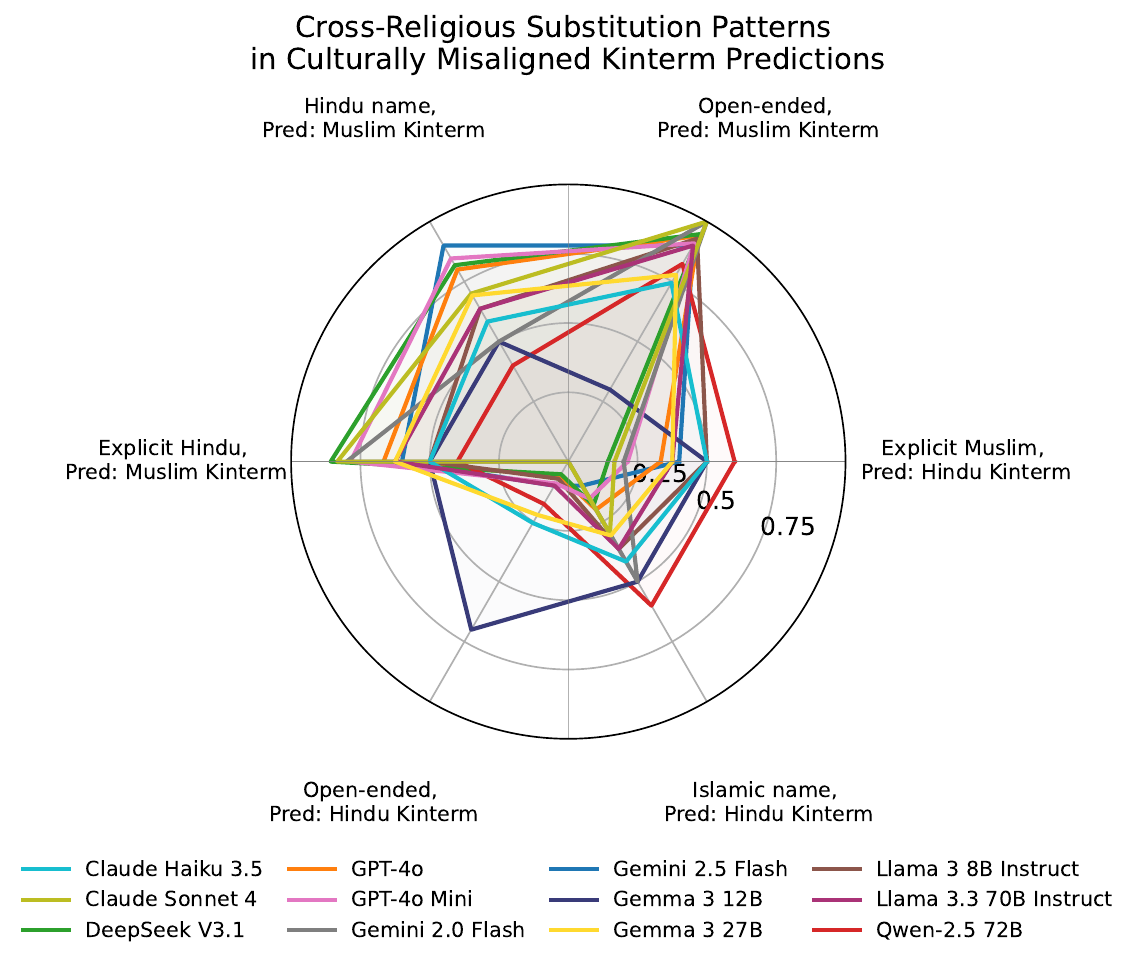}
  \caption{
  \textbf{Directional cross-religious kinship term misalignment.}
  Proportions of culturally inappropriate kinship term substitutions across explicit identity cues, implicit cues, and open-ended prompting. Misalignment is more pronounced toward substituting Muslim-associated kinterms in Hindu-marked contexts.
  }
  \label{fig:religious_misalignment}
\end{figure}

\paragraph{Language Models conflate kinship terms across religious identities}

Culturally inappropriate predictions display clear directional asymmetries across models. In all experiments, models were instructed to answer according to Bangladeshi cultural practices across multiple prompt strategies, including explicit religious identity markers (e.g., ``a Muslim person''), religiously indicative personal names (e.g., ``Arnab Roy,'' a culturally Hindu name), and open-ended prompts without overt religious cues. Several models, including GPT-4o, DeepSeek~V3.1, and Gemini~2.5~Flash, show a stronger tendency to substitute Muslim-associated kinship terms in explicitly or implicitly Hindu contexts, while fewer models exhibit the reverse pattern. To quantify this asymmetry, we perform a $\chi^2$ goodness-of-fit test for each model over all culturally inappropriate predictions involving religious mismatch. The number of such cases varies by model ($N = 24$--$65$). Eleven of twelve models exhibit statistically significant asymmetry ($p < 0.05$), with a pronounced skew toward substituting Muslim-associated kinship terms in Hindu-marked contexts. Detailed results appear in Appendix~\ref{sec:religious_misalignment}.

\begin{figure}[t]
  \centering
  \includegraphics[width=\linewidth]{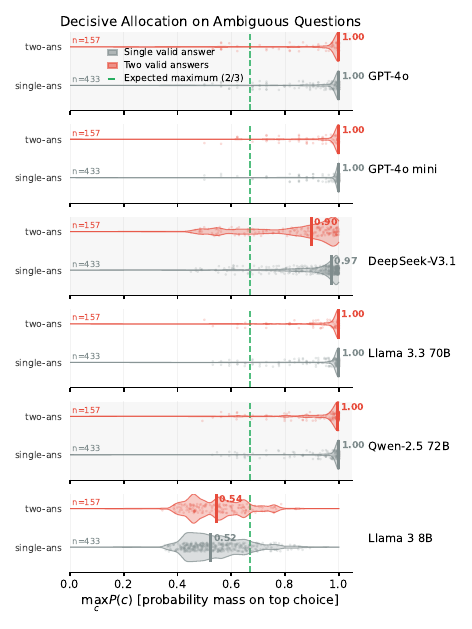}
  \caption{
    \textbf{Probability mass allocation between ambiguous and unambiguous questions.} Distribution of $\max_c P(c)$, the probability mass assigned to the most probable choice, over single-answer and two-answer questions, recovered from token-level log-probabilities via choice-restricted softmax. A well-calibrated model should show lower probability mass on two-answer questions than on single-answer ones.}
  \label{fig:logprob}
\end{figure}

\paragraph{Log-probabilities reveal decisive probability allocation on ambiguous situations} For models with accessible token-level log probabilities (Figure~\ref{fig:logprob}), we recover log-probabilities via choice-restricted softmax over pronoun logits using the top-$k$ alternatives ($k=5$) returned by the API (Appendix~\ref{logprob}), and compute $\max_c P(c)$ per question. Bangla tokenization inconsistencies were mitigated by mapping answer options to letter labels (A–-C) and restricting model outputs to these labels. Questions are split into unambiguous ($n{=}433$, single valid answer) and ambiguous ($n{=}157$, two valid answers). We use $\frac{2}{3} \approx 0.67$ as a reference boundary, above which one form holds more than twice the probability mass of any alternative. Figure~\ref{fig:logprob} shows that five of six models produce median $\max_c P(c) = 1.00$ on ambiguous questions, identical to unambiguous ones, indicating they assign near-total probability mass to a single pronoun regardless of sociopragmatic ambiguity. Only Llama~3~8B deviates (median 0.54), though still near the 0.67 boundary. This suggests that current LLMs allocate probability mass as decisively on sociopragmatically ambiguous address forms as on unambiguous ones, without reflecting the variation acknowledged by human annotators.

\section{Conclusion}
Cultural competence in language models requires not only linguistic fluency but also sensitivity to sociopragmatic context. Using \benchname, we show that even strong LLMs exhibit systematic limitations in Bangla address terms, kinship reasoning, and social customs, often defaulting to overly polite forms, lacking calibration in ambiguous contexts, and conflating kinship terminology across religious identities. Addressing the sociopragmatic weaknesses exposed by our evaluation requires both data-centric and algorithmic interventions. We hypothesize that deficits such as systematic over-politeness and cross-religious kinship conflation stem from a training data imbalance. Current multilingual LLMs are primarily exposed to formal or translated Bangla corpora, leaving informal contexts, downward-hierarchy interactions, and minority-community norms systematically underrepresented. While training native Bengali LMs from scratch may offer deeper cultural embedding, the computational requirements for general reasoning parity remain prohibitive. A more pragmatic approach lies in culturally grounded alignment. Future research could utilize culturally specific adapters or Direct Preference Optimization (DPO) with contrastive pairs derived from the failure modes identified in \benchname. Such methods would allow practitioners to leverage the reasoning capabilities of existing models while anchoring them more precisely to native Bengali relational norms.

\section*{Limitations}
While \benchname provides a focused evaluation of sociopragmatic competence in Bangladeshi social interaction, several limitations remain.

\paragraph{Standard Colloquial Bangla vs.\ dialect}
All instances and annotations are grounded in Standard Colloquial Bangla, reflecting widely shared norms of address usage and social interaction in Bangladesh. Annotators were instructed to judge cultural appropriateness with respect to this interactional standard. As a result, the benchmark does not capture the regional and dialectal variation present across Bangla varieties. Address pronouns, kinship terms, and honorifics may differ in regional varieties such as Sylheti, Chittagonian, or Rangpuri, and culturally appropriate usage in these dialects may not always align with the norms encoded in our dataset.

We restrict the benchmark to Standard Colloquial Bangla for two reasons. First, it functions as a common interactional baseline used in interregional communication, education, media, and institutional settings. Second, regional dialects differ not only lexically but also in their pragmatic conventions, which would require dialect-specific annotation guidelines and expert annotators for each variety. Without sufficient dialectal coverage, such variation could be misinterpreted as model errors.

\paragraph{Coverage of interactional contexts}
Although the dataset spans multiple sociopragmatic settings, including family, office, hospital, academic, and everyday social interactions, it cannot exhaustively represent all interactional configurations present in Bangladeshi society. Factors such as rural-–urban differences, sociolectal variation, and code-switching practices are not explicitly modeled.

Consequently, models that generate regionally appropriate but non-standard forms may be penalized under our evaluation framework. This should not be interpreted as a claim that Standard Colloquial Bangla is inherently more correct, but rather as a methodological constraint required to ensure annotation consistency and comparability across models. Future work may extend the benchmark to incorporate dialectal and regional sociopragmatic variation.

\section*{Ethical Considerations}

Annotators were compensated for their work at a rate consistent with local standards for similar annotation tasks. Prior to participation, annotators were informed about the purpose of the study and how their contributions would be used in the dataset. 

To protect privacy, annotators were instructed not to include any personally identifiable information when writing scenarios. All collected instances were additionally reviewed by the research team to ensure that they did not contain harmful, unsafe, or sensitive content. The benchmark focuses on culturally grounded everyday interactions and does not include sexual, violent, or personally identifiable material.

\bibliography{latex/custom}

\clearpage
\appendix

\section{Additional Cultural Context}
\label{sec:context}
\subsection{Bangla Address Forms}

Address forms play a central role in shaping social interaction in a language. They organize how speakers relate to one another, influence the flow of conversation, and signal identity, relationship, and context. Address forms are culturally specific, and their use connects the semantic meaning of a term with the pragmatic demands of a situation. They embody the cultural concepts, values, and social principles of a community and they help reveal how members position themselves in interaction \citep{chenChenXinRenMemeticCulturalPractice2020}. Knowledge of variation in address forms is essential in intercultural settings because it reduces the risk of sociopragmatic failure \citep{barronAcquisitionInterlanguagePragmatics2003}. To avoid the communication failure, it is essential that LLMs not only know the address terms but also understand their pragmatics. 

The formality or informality of a situation and the level of politeness or deference a speaker intends to express are reflected directly in the chosen address form \citep{ozcanChoiceAddressTerms2016}. These expressions function as linguistic markers of social and cultural identity. For language models, understanding kinship, solidarity, and power distance is especially important in Bangladeshi communication, where these relationships are strongly mirrored in conversational patterns. 

A nuanced understanding of the diversity and cultural significance of Bangla address terms is central to interpreting Bangladeshi culture. This makes it important to evaluate how well LLMs capture the cultural meanings and hierarchical cues encoded in these address forms, as address terms and honorifics together form an essential part of social deixis in the language \citep{keshavarzRoleSocialContext2001}.

\paragraph{The Pronoun of Power, Solidarity and Intimacy}

It is important to evaluate whether LLMs can recognize the social variables that shape pronominal address in Bangla, such as age, power, social distance, and intimacy. Languages differ widely in how they mark these distinctions. English relies mainly on nominal forms and lack a distinction in address pronouns \citep{dickeyFormsAddressTerms1997}, while several Asian languages use personal pronouns sparingly. Many European languages follow the T--V distinction, which separates familiar and polite forms \citep{brownPRONOUNSPOWERSOLIDARITY1968}.

Bangla offers a richer system with not two, but three second-person pronouns, apni, tumi, and tui, all meaning you but carrying different honorific and relational meanings. These forms index respect, equality, or intimacy and are closely tied to cultural expectations about how speakers negotiate hierarchy and solidarity \citep{snigdhaRepresentationSocialClass2022}. The three-way pronominal system therefore provides an informative setting for studying the sociolinguistic competence of LLMs, since correct pronoun selection requires an understanding of social hierarchy, solidarity, and interactional norms. Models that cannot infer these relational cues are likely to generate socially inappropriate or pragmatically incorrect responses.

\paragraph{Nominal Addressing in Kinship and Social Contexts}
In Bangladeshi society, kinship terms reflect relational meaning between speakers and addressees, with distinctions shaped by paternal and maternal lineage as well as religious and cultural contexts \citep{sultanaLanguageSocietyBangladesh2023,snigdhaRepresentationSocialClass2022, kolendaKinshipBangladesh1981}. The pragmatics of nominal address forms show that the social meaning of a kinship term used in interaction often differs from its literal  meaning \citep{braunTermsAddressProblems1988}. It is common in Bangladesh to address unfamiliar people with kinship-based terms that take into account age, gender, and perceived social position. Addressing elders, relatives, or even non-relatives directly by name or by surname alone is often considered disrespectful, which further reinforces the preference for kinship-based address. Because Bangladeshi society is deeply rooted in familial ties, the use of kinship terms for neighbors, acquaintances, and strangers has become a widespread and culturally expected practice.

Kinship terminology in Bangla also varies across religious communities. Muslim, Hindu, and other groups share many core kin terms but maintain distinct forms shaped by historical and ritual traditions, such as thakur-dada and thakur-ma in Hindu families, or abba-jan and amma-jan in traditional Muslim households \citep{snigdhaRepresentationSocialClass2022}. Capturing this variation is important for cultural knowledge, and our dataset therefore includes commonly used kinship terms across communities without overrepresenting any single register.

Kinship terms extend well beyond biological relations in everyday Bangladeshi interaction. They frequently function as social deictics for neighbors, acquaintances, and strangers, with speakers choosing forms like \textit{bhai} (brother), \textit{apa} (elder sister), \textit{chacha} (uncle), or \textit{nana} based on perceived age, gender, and social position. This broad extension of kinship into general social life makes kinship reasoning a crucial component of culturally appropriate communication.

\paragraph{Honorifics}
Honorifics in Bangla signal respect, politeness, and social position, and they appear in both prefix and suffix forms \citep{dasFormsAddressTerms1968}. Some are influenced by English conventions, such as Mr., Ms., and Mrs., which are used in formal and educated settings. Bangla also has a rich set of local honorifics that vary across professional, religious, and social contexts. Common suffixes include saheb, janab, and babu for men, and begum or debi for women. In workplace or service settings, speakers often use forms like sir, madam, and apa to index respect and institutional hierarchy \citep{snigdhaRepresentationSocialClass2022}.

Honorific use also overlaps with kinship-based addressing, which is widespread in Bangladeshi society. Speakers frequently use terms such as \textit{bhai}, \textit{apa}, \textit{bhabi}, \textit{chacha}, or \textit{dadu} for acquaintances or strangers based on perceived age, gender, and social position. Directly addressing elders, relatives, or even non-relatives by name alone, or by name plus surname, is often considered inappropriate, which strengthens the preference for kinship-inflected honorifics in everyday communication. These practices illustrate how Bangla honorifics encode politeness, hierarchy, and social affiliation, making them an informative component for evaluating the sociocultural competence of language models.

\subsection{Bengali Kinship Reasoning}
Bangla kinship terminology is finely differentiated and socially encoded, reflecting lineage, marriage relations, generational hierarchy, and culturally grounded expectations of respect and obligation. \citet{kolendaKinshipBangladesh1981} shows that kin terms form a core interpretive framework through which interpersonal roles and social relationships are understood. Unlike English, which relies on broad categories such as uncle, aunt, and cousin, Bangla employs a highly granular system with dozens of specific terms such as \textit{chacha} (father's brother), \textit{mama} (mother's brother), \textit{fupu} (father's sister), and \textit{dada} (paternal grandfather). This precision encodes critical relational information including lineage direction, generational rank, and relative age, making kinship not only descriptive but also a system that reinforces social hierarchy and regulates appropriate forms of respect within families and communities.

Evaluating LLMs on Bangla Kinship Reasoning is therefore valuable because correct interpretation and use of kin terms require cultural reasoning, social inference, and an understanding of relational structure that cannot be captured through grammar or vocabulary alone. A model that fails to grasp these distinctions risks producing socially inappropriate or culturally incoherent outputs.

\subsection{Bangladeshi Social Customs} 
The Social Customs domain targets culturally grounded expectations that govern appropriate behavior and response selection in everyday social interaction in Bangladesh. This subset focuses on interactional situations where multiple responses are possible, but cultural norms favor non-imposing, indirect, and harmony-preserving strategies.
Instances were constructed using a script-guided instance writing approach. Cultural scripts \citep{goddardCulturalScriptsWhat2004} were used as an abstraction layer to capture shared interactional expectations, such as softening requests, avoiding direct refusal, expressing hospitality, and managing social tension. 

\clearpage
\section{Definition and Example Instances Across Domains}
\label{sec:example_instances}
\subsection{Annotation Definitions}

\subsubsection{Pronominal Addressing}

For \textit{Pronominal Addressing}, each instance is annotated along four sociopragmatic dimensions that jointly determine appropriate second-person pronoun choice (\textit{apni}, \textit{tumi}, \textit{tui}) in Bangla.

\paragraph{Setting}
The social domain in which the interaction takes place, such as \textit{Family}, \textit{Office}, \textit{Hospital}, \textit{Academia}, or \textit{Social Interaction}. The setting constrains expected norms of formality and power relations.

\paragraph{Participant Age Relation}
The relative age hierarchy between the speaker and the addressee:
\begin{itemize}[noitemsep,leftmargin=1.2em]
    \item \textbf{Elder$\rightarrow$Younger}: the speaker is older than the addressee
    \item \textbf{Younger$\rightarrow$Elder}: the speaker is younger than the addressee
    \item \textbf{Peer$\rightarrow$Peer}: the speaker and addressee are of comparable age
\end{itemize}
Age hierarchy is a primary determinant of pronominal choice in Bangla and often overrides lexical politeness markers.

\paragraph{Participant Gender Relation}
The gender configuration of the interaction:
\begin{itemize}[noitemsep,leftmargin=1.2em]
    \item Female$\rightarrow$Male
    \item Male$\rightarrow$Female
    \item Female$\rightarrow$Female
    \item Male$\rightarrow$Male
\end{itemize}
Gender interacts with age and setting, particularly in family and institutional contexts.

\paragraph{Key}
The affective tone or manner of the interaction, following \citet{hymes1962ethnography}'s notion of \textit{Key}:
\begin{itemize}[noitemsep,leftmargin=1.2em]
    \item \textbf{Neutral}: emotionally unmarked interaction
    \item \textbf{Informal}: relaxed or intimate tone
    \item \textbf{Polite}: deferential or respectful tone
    \item \textbf{Angry}: emotionally charged or confrontational tone
\end{itemize}

\paragraph{End}
The communicative goal of the utterance:
\begin{itemize}[noitemsep,leftmargin=1.2em]
    \item \textbf{Inform}: conveying information
    \item \textbf{Inquire}: asking a question
    \item \textbf{Request}: politely seeking action or permission
    \item \textbf{Command}: issuing a directive
\end{itemize}
Correct pronominal choice requires jointly inferring these variables rather than relying on surface lexical cues.

\subsubsection{Nominal Addressing}

For \textit{Nominal Addressing}, instances are annotated to capture sociocultural variation in kinship and social titles, particularly where religious identity influences lexical choice.

\paragraph{Identity}
Whether the prompt specifies the interlocutor's religious identity:
\begin{itemize}[noitemsep,leftmargin=1.2em]
    \item \textbf{Muslim}
    \item \textbf{Hindu}
\end{itemize}

\paragraph{Type}
How identity information is conveyed in the prompt:
\begin{itemize}[noitemsep,leftmargin=1.2em]
    \item \textbf{Explicit}: religious identity is directly mentioned
    \item \textbf{Implicit}: identity is inferred through culturally indicative personal names or contextual cues
\end{itemize}

\paragraph{Basis}
The underlying reason a particular nominal form is appropriate:
\begin{itemize}[noitemsep,leftmargin=1.2em]
    \item \textbf{Religion-based}: the address term differs across religious communities
    \item \textbf{Social-based}: the term reflects social hierarchy or familiarity independent of religion
\end{itemize}

\subsubsection{Social Customs}

The \textit{Social Customs} domain evaluates culturally appropriate response selection in everyday Bangladeshi social interaction. Unlike Address Terms and Kinship Reasoning, which focus on lexical choice, this domain operationalizes culture as shared expectations about \emph{how one should respond} in common social situations.

Following the Natural Semantic Metalanguage (NSM) framework \citep{goddardCulturalScriptsWhat2004}, we identify \textbf{14 recurrent cultural scripts} that reflect widely shared Bangladeshi customs governing indirectness, politeness, emotional restraint, hospitality, and interpersonal harmony. Each script is formulated as a simple, reductive mental model using universal semantic primitives, allowing cultural expectations to be expressed without theoretical jargon and providing a structured basis for data and instance creation.
\begin{table*}[t]
\centering
\small
\begin{tabular}{c p{2.6cm} p{9.5cm}}
\toprule
\textbf{ID} & \textbf{Topic} & \textbf{Cultural Script (NSM)} \\
\midrule

1. & Request &
People think like this: when I want someone to do something, it is not good to say it strongly. If I say it strongly, this person can feel something bad. Because of this, it is good to say it softly, so this person can choose freely. \\

2. & Interrogation &
People think like this: some things are personal. Asking about these things can feel intrusive. If I ask directly, people can think I have bad manners. Because of this, it is good not to ask directly. \\

3.1 & Compliment &
People think like this: when someone says something good about me, it is not good to agree openly. People can think I think too much of myself. Because of this, it is good to say little or deny it. \\

3.2 & Compliment &
People think like this: when someone praises me too much, this person may want something from me. Because of this, it is good to respond in a way that stops more praise. \\

4. & Refusal &
People think like this: if I do not want to do something, I should not say ``no'' directly. If I do, this person can feel bad. It is better to say ``I will try'' or ``I will see.'' \\

5. & Tautology &
People think like this: some things happen not because people want them. People cannot change these things. Saying words like this helps people feel calm. \\

6. & Interjection &
People think like this: sometimes I feel something suddenly. It is good to say a small sound or word so other people know what I feel. \\

7. & Indirectness &
People think like this: when I want someone to do something, it is not good to order them. This person can feel pushed. Because of this, it is good to say it indirectly. \\

8.1 & Hospitality &
People think like this: when someone comes to my place, it is good to offer food or drink. If I do not do this, people can think badly of me. \\

8.2 & Hospitality &
People think like this: when a guest says ``I do not want to eat,'' this person says this to be polite. Because of this, it is good to insist many times. \\

9. & Greetings &
People think like this: when I see someone, it is good to say something first. Not saying anything can feel cold or bad. \\

10. & Emotion &
People think like this: when I feel something strong, it is not always good to show it in public. Other people can feel uncomfortable. Because of this, it is good to control feelings. \\

11. & Harmony &
People think like this: it is good when people feel good together. It is not good to create tension. Because of this, people speak and act gently. \\

12. & Cordiality &
People think like this: people close to me are like family. I want to be close to them, help them, and spend time with them. Distance is not good. \\

13. & Criticism &
People think like this: if someone does something wrong, it is not good to say this in front of others. This person can feel shame. It is better to say it softly or later. \\

14 & Time &
People think like this: when I say a time, it does not have to be exact. Being with people is more important than strict timekeeping. \\

\bottomrule
\end{tabular}
\caption{Cultural scripts in the Social Customs domain, expressed in Natural Semantic Metalanguage (NSM) and used as an abstraction layer to guide instance creation}
\label{tab:social_scripts}
\end{table*}

\onecolumn
\begin{table*}[t]
\centering
\large
\resizebox{\textwidth}{!}
{
\large
\begin{tabular}{c|c}
\toprule
\textbf{Subcategory} & \textbf{Example Instances} \\
\midrule

\multicolumn{2}{c}{\textbf{Address Terms: Pronominal}} \\
\midrule

\makecell[l]{
\textbf{Setting:} Family\\
\textbf{Age}: Elder$\rightarrow$Younger\\
\textbf{Gender}: Female$\rightarrow$Male\\
\textbf{Key}: Inquire\\
\textbf{End}: Angry
}
&
\makecell[l]{
\bangla{এক মা তার কিশোর ছেলেকে রাগান্বিতভাবে তার পড়াশোনার অবস্থা সম্পর্কে জিজ্ঞাসা করছেন।}\\
\textit{(A mother angrily asks her teenage son about the state of his studies.)}\\[4pt]
\bangla{এই প্রেক্ষাপটে, বাংলাদেশের শহুরে সামাজিক ব্যবস্থায় মা ছেলেকে প্রচলিত উপায়ে কীভাবে সম্বোধন করবে?}\\
\textit{(In this context, within Bangladesh’' urban social setting, how would a mother conventionally address her son?)}\\[6pt]
A.~\bangla{আপনি} (Apni)\\
B.~\bangla{তুমি} (Tumi)\\
C.~\bangla{তুই} (Tui)
} \\
\midrule

\makecell[l]{
\textbf{Setting}: Family\\
\textbf{Age}: Elder$\rightarrow$Younger\\
\textbf{Gender}: Female$\rightarrow$Female\\
\textbf{End}: Inquire\\
\textbf{Key}: Informal\\
}
&
\makecell[l]{
\bangla{এক বৃদ্ধ খালা তার শিশু ভাগ্নিকে আদুরে স্বরে প্রশ্ন করছেন।}\\
\textit{(An elderly maternal aunt affectionately asks her young niece.)}\\[4pt]
\bangla{এই প্রেক্ষাপটে, বাংলাদেশের শহুরে সামাজিক ব্যবস্থায় খালা তার ভাগ্নিকে প্রচলিত উপায়ে কীভাবে সম্বোধন করবে?}\\
\textit{(In this context, within the urban social setting of Bangladesh, how would a maternal aunt conventionally address her niece?)}\\[6pt]
A.~\bangla{আপনি} (Apni)\\
B.~\bangla{তুমি} (Tumi)\\
C.~\bangla{তুই} (Tui)
} \\

\midrule

\makecell[l]{
\textbf{Setting}: Office\\
\textbf{Age}: Younger$\rightarrow$Elder\\
\textbf{Gender}: Male$\rightarrow$Male\\
\textbf{End}: Request\\
\textbf{Key}: Polite\\
}
&
\makecell[l]{
\bangla{একজন মধ্যবয়স্ক সহকারী সচিব তার বয়স্ক উপসচিবকে সরকারি তহবিল ব্যবহারে বিনীত অনুরোধ করছেন।}\\
\textit{A middle-aged Assistant Secretary is politely requesting an older Deputy Secretary regarding the use of government funds.)}\\[4pt]
\bangla{এই প্রেক্ষাপটে, বাংলাদেশী সংস্কৃতিতে সহকারী সচিব উপসচিবকে প্রচলিত উপায়ে কীভাবে সম্বোধন করবেন?}\\
\textit{(In this context, within Bangladeshi culture, how would the Assistant Secretary conventionally address the Deputy Secretary?)}\\[6pt]
A.~\bangla{আপনি} (Apni)\\
B.~\bangla{তুমি} (Tumi)\\
C.~\bangla{তুই} (Tui)
} \\

\midrule

\makecell[l]{
\textbf{Setting}: Office\\
\textbf{Age}: Elder$\rightarrow$Younger\\
\textbf{Gender}: Female$\rightarrow$Female\\
\textbf{End}: Inform\\
\textbf{Key}: Angry\\
}
&
\makecell[l]{
\bangla{একজন বয়স্ক নারী উপ-পরিচালক তার মধ্যবয়স্ক নারী সহকারি পরিচালককে পাসপোর্ট যাচাইকরণের নতুন নিয়মাবলী সম্পর্কে রাগান্বিত হয়ে অবহিত করছেন।}\\
\makecell[l]{\textit{(An elderly female Deputy Director is angrily informing her middle-aged female Assistant Director about} \\ \textit{the new regulations for passport verification.)}}\\[4pt]
\bangla{এই প্রেক্ষাপটে, বাংলাদেশী সংস্কৃতিতে উপ-পরিচালক সহকারি পরিচালককে প্রচলিত উপায় কিভাবে সম্বোধন করবেন?}\\
\textit{(In this context, within Bangladeshi culture, how would the Deputy Director conventionally address the Assistant Director?)}\\[6pt]
A.~\bangla{আপনি} (Apni)\\
B.~\bangla{তুমি} (Tumi)\\
C.~\bangla{তুই} (Tui)
} \\

\midrule

\makecell[l]{
\textbf{Setting}: Academia\\
\textbf{Age}: Younger$\rightarrow$Elder\\
\textbf{Gender}: Male$\rightarrow$Female\\
\textbf{End}: Request\\
\textbf{Key}: Polite\\
}
&
\makecell[l]{
\bangla{স্কুলের একজন দ্বিতীয় শ্রেণীর ছাত্র বিনয়ের সাথে তার গণিত শিক্ষিকাকে অংকটি দেখতে অনুরোধ করল।}\\
\textit{(A second-grade student politely asks his mathematics teacher to look at a problem.)}\\[4pt]
\bangla{এই প্রেক্ষাপটে, বাংলাদেশী শহুরে সামাজিক ব্যবস্থায় দ্বিতীয় শ্রেণীর ছাত্রটি তার গণিত শিক্ষিকাকে প্রচলিত উপায় কিভাবে সম্বোধন করবে?}\\
\makecell[l]{\textit{(In this context, within Bangladesh’'s urban social setting, how would the second-grade student conventionally address}\\ \textit{his mathematics teacher?)}}\\[6pt]
A.~\bangla{আপনি} (Apni)\\
B.~\bangla{তুমি} (Tumi)\\
C.~\bangla{তুই} (Tui)
} \\

\midrule

\makecell[l]{
\textbf{Setting}: Academia\\
\textbf{Age}: Peer$\rightarrow$Peer\\
\textbf{Gender}: Male$\rightarrow$Female\\
\textbf{End}: Command\\
\textbf{Key}: Angry\\
}
&
\makecell[l]{
\bangla{পঞ্চম শ্রেণীর একজন ছাত্র তার একই ক্লাসের বান্ধবীকে রাগান্বিত স্বরে ডাকছে।}\\
\textit{(A fifth-grade male student is angrily addressing a female classmate from the same class.)}\\[4pt]
\bangla{এই প্রেক্ষাপটে, বাংলাদেশী শহুরে সামাজিক ব্যবস্থায় তারা একে অন্যকে প্রচলিত উপায় কিভাবে সম্বোধন করবে?}\\
\textit{(In this context, within Bangladesh’'s urban social setting, how would the two conventionally address each other?)}\\[6pt]
A.~\bangla{আপনি} (Apni)\\
B.~\bangla{তুমি} (Tumi)\\
C.~\bangla{তুই} (Tui)
} \\
\midrule

\makecell[l]{
\textbf{Setting}: Hospital\\
\textbf{Age}: Younger$\rightarrow$Elder\\
\textbf{Gender}: Female$\rightarrow$Male\\
\textbf{End}: Inquire\\
\textbf{Key}: Neutral\\
}
&
\makecell[l]{
\bangla{একজন তরুণ বয়সী মহিলা স্টাফ নার্স মধ্যবয়স্ক পুরুষ ইমার্জেন্সি মেডিকেল অফিসারকে ঔষধের মাত্রা নিয়ে জিজ্ঞাসা করল।}\\
\textit{(A young female staff nurse asks a middle-aged male emergency medical officer about the medication dosage.)}\\[4pt]
\bangla{এই প্রেক্ষাপটে, বাংলাদেশী সংস্কৃতিতে স্টাফ নার্স ইমার্জেন্সি মেডিকেল অফিসারকে প্রচলিত উপায়ে কীভাবে সম্বোধন করবেন?}\\
\textit{(In this context, within Bangladeshi culture, how would the staff nurse conventionally address the emergency medical officer?)}\\[6pt]
A.~\bangla{আপনি} (Apni)\\
B.~\bangla{তুমি} (Tumi)\\
C.~\bangla{তুই} (Tui)
} \\

\midrule

\makecell[l]{
\textbf{Setting}: Social Interaction\\
\textbf{Age}: Elder$\rightarrow$Younger\\
\textbf{Gender}: Female$\rightarrow$Female\\
\textbf{End}: Inform\\
\textbf{Key}: Informal\\
}
&
\makecell[l]{
\bangla{আফরিনের বাসায় প্রতিবেশীর ছোট একটি মেয়ে শিশু বেড়াতে আসলে আফরিন তাকে আদর করে স্কুলের মজার ঘটনার বিবরণ দিচ্ছে।}\\
\textit{(When a young girl from a neighboring family visits Afrin’s home, Afrin affectionately tells her about a funny incident from school.)}\\[4pt]
\bangla{এই প্রেক্ষাপটে, বাংলাদেশের শহুরে সামাজিক ব্যবস্থায় আফরিন শিশুটিকে প্রচলিত উপায়ে কীভাবে সম্বোধন করবে?}\\
\textit{(In this context, within an urban Bangladeshi social setting, how would Afrin typically address the child?)}\\[6pt]
A.~\bangla{আপনি} (Apni)\\
B.~\bangla{তুমি} (Tumi)\\
C.~\bangla{তুই} (Tui)
} \\

\midrule

\makecell[l]{
\textbf{Setting}: Social Interaction\\
\textbf{Age}: Younger$\rightarrow$Elder\\
\textbf{Gender}: Male$\rightarrow$Male\\
\textbf{End}: Command\\
\textbf{Key}: Neutral\\
}
&
\makecell[l]{
\bangla{একজন কিশোর বয়সী ছেলে বয়স্ক মুচিকে জুতা পরিষ্কার করতে বলছে।}\\
\textit{(A teenage boy is asking an elderly cobbler to clean his shoes.)}\\[4pt]
\bangla{এই প্রেক্ষাপটে, বাংলাদেশের শহুরে সামাজিক ব্যবস্থায় কিশোর মুচিকে প্রচলিত উপায়ে কীভাবে সম্বোধন করবে?}\\
\textit{(In this context, within an urban Bangladeshi social setting, how would the teenager typically address the cobbler?)}\\[6pt]
A.~\bangla{আপনি} (Apni)\\
B.~\bangla{তুমি} (Tumi)\\
C.~\bangla{তুই} (Tui)
} \\
\bottomrule
\end{tabular}
}
\caption{Example instances from the Pronominal Addressing subset of \textit{Address Terms} domain. Each instance is further annotated with sociopragmatic metadata, namely Setting, Participant Age, Participant Gender, Key, and End.}
\label{tab:example_overview_pronominal}
\end{table*}

\clearpage

\begin{table*}[t]
\centering
\large
\resizebox{\textwidth}{!}
{
\large
\begin{tabular}{c|c}
\toprule
\textbf{Subcategory} & \textbf{Example Instances} \\
\midrule

\multicolumn{2}{c}{\textbf{Address Terms: Nominal}} \\
\midrule

\makecell[l]{
\textbf{Subtopic:} Kinship\\
\textbf{Identity}: Muslim\\
\textbf{Type}: Implicit\\
\textbf{Basis}: Religion\\
}
&
\makecell[l]{
\bangla{বাংলাদেশী সংস্কৃতি অনুযায়ী কবীর তার মায়ের বোনকে প্রচলিত উপায়ে সম্বোধন করবে?}\\
\textit{(According to Bangladeshi culture, how would Kabir conventionally address his mother's sister?)}\\[6pt]
A.~\bangla{খালা} \textit{(Khala)}\\
B.~\bangla{ফুপু} \textit{(Fupu)}\\
C.~\bangla{মাসি} \textit{(Mashi)}\\
D.~\bangla{পিসি} \textit{(Pishi)}\\
} \\
\midrule

\makecell[l]{
\textbf{Subtopic:} Kinship\\
\textbf{Identity}: Hindu\\
\textbf{Type}: Implicit\\
\textbf{Basis}: Religion\\
}
&
\makecell[l]{
\bangla{বাংলাদেশী সংস্কৃতি অনুযায়ী প্রণব রায় তার মায়ের মাকে প্রচলিত উপায়ে সম্বোধন কুরবে?}\\
\textit{(According to Bangladeshi culture, how would Pranab Roy conventionally address his mother's mother?)}\\[6pt]
A.~\bangla{ঠাকুরমা} \textit{(Thakurma)}\\
B.~\bangla{দাদি} \textit{(Dadi)}\\
C.~\bangla{জেঠী} \textit{(Jethi)}\\
D.~\bangla{নানি} \textit{(Nani)}
} \\

\midrule

\makecell[l]{
\textbf{Subtopic:} Kinship\\
\textbf{Identity}: Muslim\\
\textbf{Type}: Implicit\\
\textbf{Basis}: Neutral\\
}
&
\makecell[l]{
\bangla{বাংলাদেশী সংস্কৃতি অনুযায়ী সিদ্দিকের ভাই এর মেয়ে সম্পর্কে তার কে হয়?}\\
\textit{(According to Bangladeshi culture, what is Siddik's relationship to his brother's daughter?)}\\[6pt]
A.~\bangla{ভাইঝি} \textit{(Bhaijhi)}\\
B.~\bangla{ভাগনি} \textit{(Bhagni)}\\
C.~\bangla{নাতনি} \textit{(Natni)}\\
D.~\bangla{দৌহিত্রী} \textit{(Douhitri)}
} \\

\midrule

\makecell[l]{
\textbf{Subtopic:} Social Addressing\\
\textbf{Identity}: Hindu\\
\textbf{Type}: Implicit\\
\textbf{Basis}: Religion\\
}
&
\makecell[l]{
\bangla{ফ্ল্যাটে ঢোকার সময় একজন তরুণ বয়সী ভাড়াটিয়া নির্মল দে মধ্যবয়সী দারোয়ান কানাই নাথকে দরজা খোলার কথা বলছেন}\\
\textit{(While entering the apartment, a young tenant, Nirmal Dey, is asking the middle-aged doorman, Kanai Nath, to open the door.)}\\[4pt]
\bangla{বাংলাদেশী সংস্কৃতিতে ভাড়াটিয়া দাড়োয়ানকে প্রচলিত উপায়ে কীভাবে সম্বোধন করবে?}\\
\textit{(In this context, according to Bangladeshi culture, how would the tenant conventionally address the doorman?)}\\[6pt]
A.~\bangla{ভাই} \textit{(Bhaijhi)}\\
B.~\bangla{দাদা} \textit{(Dada)}\\
C.~\bangla{কানাই} \textit{(Kanai)}\\
D.~\bangla{পিসে} \textit{(Pise)}
} \\

\midrule

\makecell[l]{
\textbf{Subtopic:} Social Addressing\\
\textbf{Identity}: Hindu\\
\textbf{Type}: Explicit\\
\textbf{Basis}: Religion\\
}
&
\makecell[l]{
\bangla{একজন মধ্যবয়সী হিন্দু মহিলা উর্মি রায় মধ্যবয়সী হিন্দু দর্জি প্রণব কুমারকে কাপড় কবে পাওয়া যাবে জিজ্ঞেস করছে।}\\
\textit{(A middle-aged Hindu woman, Urmi Roy, is asking a middle-aged Hindu tailor, Pranab Kumar, when the clothes will be ready.)}\\[4pt]
\bangla{এই প্রেক্ষাপটে, বাংলাদেশী সংস্কৃতিতে মহিলা দর্জিকে প্রচলিত উপায়ে কীভাবে সম্বোধন করবে?}\\
\textit{(In this context, according to Bangladeshi culture, how would the woman conventionally address the tailor?)}\\[6pt]
A.~\bangla{দাদা} \textit{(Dada)}\\
B.~\bangla{ভাই} \textit{(Bhai)}\\
C.~\bangla{মামা} \textit{(Mama)}\\
D.~\bangla{চাচা} \textit{(Chacha)}
} \\

\midrule

\makecell[l]{
\textbf{Subtopic:} Social Addressing\\
\textbf{Identity}: Neutral\\
\textbf{Type}: Neutral\\
\textbf{Basis}: Social\\
}
&
\makecell[l]{
\bangla{ট্রেন প্ল্যাটফর্মে একজন কিশোর একজন বৃদ্ধ অপরিচিতকে লাইনের শেষে দাঁড়াতে অনুরোধ করছে।}\\
\textit{(On a train platform, a teenager is asking an elderly stranger to stand at the end of the line.)}\\[4pt]
\bangla{এই প্রেক্ষাপটে, বাংলাদেশী সামাজিক ব্যবস্থায় কিশোরটি বৃদ্ধ ব্যক্তিকে প্রচলিত উপায়ে কীভাবে সম্বোধন করবে?}\\
\textit{(In this context, according to Bangladeshi culture, how would the woman conventionally address the tailor?)}\\[6pt]
A.~\bangla{চাচা} \textit{(Chacha)}\\
B.~\bangla{ভাই} \textit{(Bhai)}\\
C.~\bangla{মামা} \textit{(Mama)}\\
D.~\bangla{দাদা} \textit{(Dada)}
} \\

\midrule

\makecell[l]{
\textbf{Subtopic:} Honorifics\\
\textbf{Identity}: Neutral\\
\textbf{Type}: Neutral\\
\textbf{Basis}: Neutral\\
}
&
\makecell[l]{
\bangla{এক নারী মহিলা উপজেলা নির্বাহী কর্মকর্তার সঙ্গে কথা বলছে।}\\
\textit{(A woman is speaking with a female Upazila Nirbahi Officer (UNO).)}\\[4pt]
\bangla{এই প্রেক্ষাপটে বাংলাদেশী সংস্কৃতিতে কোন সম্মানসূচক সম্বোধনটি সবচেয়ে উপযুক্ত?}\\
\textit{(In this context, according to Bangladeshi culture, which honorific form of address is most appropriate?)}\\[6pt]
A.~\bangla{ম্যাডাম} \textit{(Madam)}\\
B.~\bangla{মিস} \textit{(Miss)}\\
C.~\bangla{মিসেস} \textit{(Mrs)}\\
D.~\bangla{বেগম} \textit{(Begum)}
} \\
\midrule

\makecell[l]{
\textbf{Subtopic:} Honorifics\\
\textbf{Identity}: Hindu\\
\textbf{Type}: Implicit\\
\textbf{Basis}: Religion\\
}
&
\makecell[l]{
\bangla{একই গুরুর কাছে দীক্ষা নেওয়া পুরুষ শিষ্যেরা একে অপরের সাথে কথা বলছে}\\
\textit{Male disciples initiated by the same guru are talking to each other.)}\\[4pt]
\bangla{এই প্রেক্ষাপটে বাংলাদেশী সংস্কৃতিতে কোন সম্মানসূচক সম্বোধনটি সবচেয়ে উপযুক্ত?}\\
\textit{(In this context, according to Bangladeshi culture, which honorific form of address is most appropriate?)}\\[6pt]
A.~\bangla{গুরুভাই} \textit{(Gurubhai)}\\
B.~\bangla{সতীর্থ} \textit{(Shotirtho)}\\
C.~\bangla{দাদা} \textit{(Dada)}\\
D.~\bangla{বন্ধু} \textit{(Friend)}
} \\
\bottomrule
\end{tabular}
}
\caption{Example instances from the Nominal Addressing subset of \textit{Address Terms} domain. Each instance is further annotated with Identity, Type, Basis.}
\label{tab:overview_nominal}
\end{table*}

\clearpage
\begin{table*}[t]
\centering
\large
\resizebox{\textwidth}{!}
{
\large
\begin{tabular}{c|c}
\toprule
\textbf{Subcategory} & \textbf{Example Instances} \\
\midrule

\multicolumn{2}{c}{\textbf{Kinship Reasoning}} \\
\midrule

\makecell[l]{
\textbf{Hop Count:} 2\\
}
&
\makecell[l]{
\bangla{রেশমা-এর মাতা নার্গিস। তনিম-নার্গিস এর কন্যা নয়, তবে রেশমার ভাই হয়। অতএব, তনিম সম্পর্কে নার্গিসের কে হয়?}\\
\textit{(Nargis is Reshma's mother. Tanim is not Nargis's daughter, but Reshma's brother. Therefore, who is Tanim to Nargis?)}\\[6pt]
A.~\bangla{ছেলে} \textit{(Son)}\\
B.~\bangla{ভাই} \textit{(Brother)}\\
C.~\bangla{দাদা} \textit{(Paternal Grandfather)}\\
D.~\bangla{বাবা} \textit{(Father)}\\
} \\
\midrule

\makecell[l]{
\textbf{Hop Count:} 3\\
}
&
\makecell[l]{
\bangla{একটি মহিলাকে নির্দেশ করে, অপর একটি মহিলা বললেন “তার ভাইয়ের বাবা হলেন আমার দাদার একমাত্র ছেলে। মহিলাটি অপর মহিলার কে হন?}\\
\textit{(Referring to a woman, another woman said, "Her brother's father is my grandfather's only son."
What is the relationship between the two women?)}\\[6pt]
A.~\bangla{বোন} \textit{(Sister)}\\
B.~\bangla{মা} \textit{(Mother)}\\
C.~\bangla{চাচি} \textit{(Aunt)}\\
D.~\bangla{মেয়ে} \textit{(Daughter)}
} \\

\midrule

\makecell[l]{
\textbf{Hop Count:} 4\\
}
&
\makecell[l]{
\bangla{আকাশের মা হলেন সালমা, যিনি আবার কামালের বোন। কামালের মেয়ের নাম রিয়া এবং রিয়া হলো পলাশের স্ত্রী। এদিকে, রফিক হলেন সালমার স্বামী। তাহলে, রফিক আকাশের সম্পর্কে কে হন?}\\
\makecell[l]{\textit{(Akash's mother is Salma, who is also Kamal's sister. Kamal's daughter is named Riya, and Riya is Palash's wife. Meanwhile, Rafiq is Salma's husband.} \\  \textit{What is Rafiq's relationship to Akash?)}}\\[6pt]
A.~\bangla{বাবা} \textit{(Father)}\\
B.~\bangla{কাকা} \textit{(Paternal uncle)}\\
C.~\bangla{স্বামী} \textit{(Husband)}\\
D.~\bangla{পুত্র} \textit{(Son)}
} \\

\midrule

\makecell[l]{
\textbf{Hop Count:} 4\\
}
&
\makecell[l]{
\bangla{রাশেদের বাবার বোনের বাবার ছেলের মেয়ে আমার সম্পর্কে রাশেদের কে হন?}\\
\textit{(Who is the daughter of the son of the father of Rashed's father's sister to Rashed?)}\\[6pt]
A.~\bangla{বোন} \textit{(Sister)}\\
B.~\bangla{খালা} \textit{(Maternal aunt)}\\
C.~\bangla{ফুফু} \textit{(Paternal aunt)}\\
D.~\bangla{ভাবি} \textit{(Sister-in-law)}
} \\

\midrule

\makecell[l]{
\textbf{Hop Count:} 1\\
}
&
\makecell[l]{
\bangla{আরমান, বেলাল ও রাশেদ তিন ভাই। রাশেদ হলো ইমরানের বাবা এবং ইমরান হলো সুমনের ভাই। তাহলে আরমান সম্পর্কে রাশেদের কে হন?}\\
\textit{(Arman, Belal, and Rashed are three brothers. Rashed is Imran's father, and Imran is Sumon's brother. Then, who is Arman to Rashed?)}\\[6pt]
A.~\bangla{চাচা} \textit{(Paternal Uncle)}\\
B.~\bangla{ভাতিজা} \textit{(Nephew)}\\
C.~\bangla{দাদা} \textit{(Paternal Grandfather)}\\
D.~\bangla{মামা} \textit{(Maternal Uncle)}
} \\

\bottomrule
\end{tabular}
}
\caption{Example instances from the Kinship Reasoning Domain. Instances involve kinship chains with hop counts ranging from one to four.}
\label{tab:example_prompt_kin_reasoning}
\end{table*}

\clearpage

\begin{table*}[t]
\centering
\large
\resizebox{\textwidth}{!}
{
\large
\begin{tabular}{c|c}
\toprule
\textbf{Subcategory} & \textbf{Example Instances} \\
\midrule

\multicolumn{2}{c}{\textbf{Social Customs}} \\
\midrule

\makecell[l]{
\textbf{Request}\\
}
&
\makecell[l]{
\bangla{একজন যাত্রী মনে করছে রিকশাভাড়া একটু বেশি বলা হয়েছে। তিনি ভাড়া কমানোর জন্য রিকশাওয়ালার সাথে কথা বলছেন}\\
\textit{(A passenger thinks that the rickshaw fare quoted is a bit high and is talking to the rickshaw puller to negotiate a lower fare.)}\\[4pt]
\bangla{এই প্রেক্ষাপটে বাংলাদেশী সংস্কৃতিতে সবচেয়ে প্রচলিত প্রকাশভঙ্গী কোনটি?}\\
\textit{(In this context, in Bangladeshi culture, which expression is the most commonly used?)}\\[6pt]
A.~\bangla{ভাড়া বেশি বলছো} \textit{(You're asking for too much fare)}\\
B.~\bangla{এই ভাড়ায় কেউ যাবে না} \textit{(No one would go at this fare.)}\\
C.~\bangla{দয়া করে ভাড়া কমাও} \textit{(Please lower the fare.)}\\
D.~\bangla{একটু কম হলে ভালো হতো} \textit{(It would be better if it were a bit less.)}\\
} \\
\midrule

\makecell[l]{
\textbf{Interrogation}\\
}
&
\makecell[l]{
\bangla{মাসের শেষ দিনে টিউশনে গিয়ে দেখলেন অভিভাবক স্যালারি দিতে ভুলে গেছেন}\\
\textit{(On the last day of the month, you go to a tutoring session and realize that the guardian has forgotten to pay your salary.)}\\[4pt]
\bangla{এই প্রেক্ষাপটে বাংলাদেশী সংস্কৃতিতে আপনার জন্য সবচেয়ে উপযুক্ত কোনটি?}\\
\textit{(In this context, in Bangladeshi culture, which option would be most appropriate for you?)}\\[6pt]
A.~\bangla{প্রসঙ্গ এর জন্য অপেক্ষা করবেন, এবং সুযোগ পেলে অভিভাবককে বিষয়টা জানাবেন}\\ 
\textit{(Wait for an appropriate moment and inform the guardian when you get the chance.)}\\
B.~\bangla{সরাসরি না বলে ইশারা করবেন} \textit{(Hint at it indirectly instead of saying it directly)}\\
C.~\bangla{রেগে গিয়ে অভিভাবককে ডাকবেন} \textit{(Get angry and call out the guardian)}\\
D.~\bangla{সরাসরি বলবেন, "স্যালারি দিন"} \textit{(Say directly, "Give me my salary.")}\\
} \\

\midrule

\makecell[l]{
\textbf{Compliment}\\
}
&
\makecell[l]{
\bangla{শিক্ষক বললেন, “তোমার উত্তরটা ক্লাসে সবচেয়ে ভালো ছিল।”}\\
\textit{(The teacher said, "Your answer was the best in the class.')}\\[4pt]
\bangla{এই প্রেক্ষাপটে বাংলাদেশী সংস্কৃতিতে সবচেয়ে উপযুক্ত উত্তর কোনটি?}\\
\textit{(In this context, according to Bangladeshi culture, which response would be the most appropriate?)}\\[6pt]
A.~\bangla{ধন্যবাদ, আমি খুব ভালো} \textit{(Thank you, I am very good.)}\\
B.~\bangla{পাশের বন্ধু আমাকে শিখিয়ে দিয়েছে, তাই পেরেছি} \textit{(A friend sitting next to me helped me, so I managed it.)}\\
C.~\bangla{আপনার জন্যই তো পারলাম} \textit{(I was able to do it because of you)}\\
D.~\bangla{আমি সবসময়ই ভালো করি} \textit{(I always do well.)}
} \\

\midrule

\makecell[l]{
\textbf{Refusal} \\
}
&
\makecell[l]{
\bangla{একজন আত্মীয় আপনাকে আগামীকাল গ্রামের বাড়িতে যেতে বলছেন, কিন্তু আপনি যেতে চান না।}\\
\textit{(A relative is asking you to go to the village home tomorrow, but you do not want to go.”)}\\[4pt]
\bangla{এই প্রেক্ষাপটে বাংলাদেশী সংস্কৃতিতে সবচেয়ে উপযুক্ত উত্তর কোনটি?}\\
\textit{(In this context, according to Bangladeshi culture, which response would be the most appropriate?)}\\[6pt]
A.~\bangla{সময়-সুযোগ হলে অবশ্যই যাব} \textit{(I will definitely go when time and circumstances allow.)}\\
B.~\bangla{আমার এখন একদম ইচ্ছা নেই} \textit{(I really don't feel like going right now.)}\\
C.~\bangla{চেষ্টা করব কালই যাওয়ার} \textit{(I will try to go tomorrow.)}\\
D.~\bangla{যাব না} \textit{(I will not go.)}
} \\

\midrule

\makecell[l]{
\textbf{Tautology} \\
}
&
\makecell[l]{
\bangla{অনেক চেষ্টা করেও একজন ছাত্র বিশ্ববিদ্যালয় ভর্তি পরীক্ষায় পাস করতে পারেনি বলে তাকে সান্ত্বনা দিচ্ছেন}\\
\textit{(After many attempts, a student failed to pass the university admission exam, and you are consoling him.)}\\[4pt]
\bangla{এই প্রেক্ষাপটে বাংলাদেশী সংস্কৃতিতে সবচেয়ে উপযুক্ত উত্তর কোনটি?}\\
\textit{(In this context, according to Bangladeshi culture, which response would be the most appropriate?)}\\[6pt]
A.~\bangla{ভালো ভাবে পড়লে এমন হত না} \textit{(If you had studied properly, this wouldn't have happened.)}\\
B.~\bangla{এতেই বরং ভবিষ্যতে ভালো হবে} \textit{(This will actually be better for the future.)}\\
C.~\bangla{আর একটু চেষ্টা করলে হতো} \textit{(If you had tried a little more, you would have succeeded.)}\\
D.~\bangla{আল্লাহ যা করেন মঙ্গলের জন্যই করেন, হয়তো কপালে এর চেয়ে ভালো কিছু আছে} \\ 
\textit{(Whatever Allah does is for the best; perhaps something better than this is destined for you.)}
} \\

\midrule

\makecell[l]{
\textbf{Interjection} \\
}
&
\makecell[l]{
\bangla{রাস্তায় হাঁটার সময় হঠাৎ পায়ে হোঁচট খেলেন}\\
\textit{(While walking on the road, you suddenly stumble.)}\\[4pt]
\bangla{এই প্রেক্ষাপটে বাংলাদেশী সংস্কৃতিতে তাৎক্ষণিকভাবে কোন আবেগ প্রকাশক ধ্বনিটি প্রচলিত?}\\
\textit{(In this context, according to Bangladeshi culture, which immediate emotional exclamation is commonly used?)}\\[6pt]
A.~\bangla{ছিঃ!} \textit{(Chi!)}\\
B.~\bangla{ধুর!} \textit{(Dhur!)}\\
C.~\bangla{আহা!} \textit{(Aha!)}\\
D.~\bangla{উফ!} \textit{(Uf!)}
} \\

\midrule

\makecell[l]{
\textbf{Indirectness} \\
}
&
\makecell[l]{
\bangla{মিটিংয়ে বয়স্ক সহকর্মী একটি প্ল্যান দিলেন যা আপনার কাছে অবাস্তব মনে হচ্ছে।}\\
\textit{(In a meeting, a colleague proposes a plan that seems unrealistic to you.
)}\\[4pt]
\bangla{এই প্রেক্ষাপটে বাংলাদেশী সংস্কৃতিতে সবচেয়ে প্রচলিত প্রকাশভঙ্গী কোনটি?}\\
\textit{(In this context, according to Bangladeshi culture, which expression is most commonly appropriate?)}\\[6pt]
A.~\bangla{আমি আপনার আইডিয়ার সাথে একমত হতে পারছি না কারণ এতে লজিক্যাল গ্যাপ আছে} \\ 
\textit{(I cannot agree with your idea because it has a logical gap.)}\\
B.~\bangla{আইডিয়াটা ইন্টারেস্টিং, তবে আমাদের বর্তমান রিসোর্সের সাথে এটা কতটা ম্যাচ করবে তা একটু ভেবে দেখা দরকার}\\
\textit{(The idea is interesting, but we should think a bit about how well it matches our current resources.)}\\
C.~\bangla{আপনি কি এই আইডিয়াটা সকালে ভেবেছেন নাকি রাতে?}\\
\textit{(Did you think of this idea in the morning or at night?)}\\
D.~\bangla{এটা খুবই ফালতু একটা আইডিয়া, এটা কাজ করবে না} \\
\textit{(This is a very useless idea; it will not work.)}
} \\

\midrule

\bottomrule
\end{tabular}
}
\caption{Example instances from the Social Customs Domain. Contd..}
\label{tab:example_prompt_customs_1}
\end{table*}

\clearpage

\begin{table*}[t]
\centering
\large
\resizebox{\textwidth}{!}
{
\large
\begin{tabular}{c|c}
\toprule
\textbf{Subcategory} & \textbf{Example Instances} \\
\midrule

\multicolumn{2}{c}{\textbf{Social Customs}} \\
\midrule

\makecell[l]{
\textbf{Hospitality}\\
}
&
\makecell[l]{
\bangla{মেহমানকে খাবার দিলে তিনি ভদ্রতা করে বলছেন, "না না, কিছু খাব না"।}\\
\textit{(When you offer food to a guest, they politely say, "No, no, I won't eat anything.")}\\[4pt]
\bangla{এই প্রেক্ষাপটে বাংলাদেশী সংস্কৃতিতে সবচেয়ে প্রচলিত উপায়ে আপনি কী করবেন?}\\
\textit{(In this context, according to Bangladeshi culture, what would you most commonly do?)}\\[6pt]
A.~\bangla{কিছুই দেবেন না} \textit{(Not offer anything)}\\
B.~\bangla{খাবার নষ্ট হবে না ভেবে খুশি হবেন} \textit{(Feel happy thinking that the food will not go to waste)}\\
C.~\bangla{শুধু পানি দিবেন} \textit{(Offer only water.)}\\
D.~\bangla{জোর করে বলবেন, 'সামান্য একটু মুখে দিন, না খেলে আমি কষ্ট পাব।'}\\
\textit{(Insist by saying, "Please have just a little; I will feel hurt if you don't eat.")}\\
} \\
\midrule

\makecell[l]{
\textbf{Greetings}\\
}
&
\makecell[l]{
\bangla{বন্ধুর বাবা-মাকে রাস্তায় দেখলেন।}\\
\textit{(You see your friend's parents on the street.)}\\[4pt]
\bangla{এই প্রেক্ষাপটে বাংলাদেশী সংস্কৃতিতে সবচেয়ে প্রচলিত উপায়ে আপনি কী করবেন?}\\
\textit{(In this context, according to Bangladeshi culture, what would be the most common thing to do?)}\\[6pt]
A.~\bangla{হাসবেন} \textit{(Smile)}\\
B.~\bangla{সালাম দিয়ে বলবেন, আঙ্কেল/আন্টি শরীর ভালো?}\\ 
\textit{(Greet them with salaam and say, "Uncle/Auntie, how are you?")}\\
C.~\bangla{মাথা নিচু করে থাকবেন} \textit{(Keep your head lowered.)}\\
D.~\bangla{চুপ থাকবেন} \textit{(Remain silent.)}\\
} \\

\midrule

\makecell[l]{
\textbf{Emotion}\\
}
&
\makecell[l]{
\bangla{মার্কেটে কেনাকাটা করার সময় স্ত্রীর/স্বামীর কোনো আচরণে প্রচণ্ড রাগ হলো}\\
\textit{(While shopping at the market, you become extremely angry at some behavior of your wife/husband.)}\\[4pt]
\bangla{এই প্রেক্ষাপটে বাংলাদেশী সংস্কৃতিতে সবচেয়ে প্রচলিত উপায়ে আপনি কী করবেন?}\\
\textit{(In this context, according to Bangladeshi culture, what would be the most common response?)}\\[6pt]
A.~\bangla{সবার সামনে কিছু না বলে স্বাভাবিক থাকবেন, বাসায় ফিরে কথা বলবেন)}\\\textit{(Stay normal in public without saying anything, and discuss it after returning home.)}\\
B.~\bangla{ইঙ্গিতে বলবেন, "আমি এখন আমার ক্রোধের মাত্রা ১০-এর স্কেলে ৮ অনুভব করছি"}\\
\textit{(Indirectly say, "I am currently feeling my anger at an 8 on a scale of 10.")}\\
C.~\bangla{সবার সামনে তাকে ধাক্কা দিয়ে বা অপমান করে চলে আসবেন} \textit{(Push or insult them in public and leave.)}\\
D.~\bangla{দোকানের মধ্যেই চিৎকার করে বলবেন যে তার আচরণ আপনার পছন্দ হচ্ছে না}\\ \textit{(Shout inside the shop that you do not like their behavior.)}
} \\

\midrule

\makecell[l]{
\textbf{Harmony} \\
}
&
\makecell[l]{
\bangla{ঈদের দিনে হিন্দু প্রতিবেশীর সাথে দেখা।}\\
\textit{(You meet your Hindu neighbor on the day of Eid.
)}\\[4pt]
\bangla{এই প্রেক্ষাপটে বাংলাদেশী সংস্কৃতিতে সবচেয়ে প্রচলিত উপায়ে আপনি কী করবেন?}\\
\textit{(In this context, according to Bangladeshi culture, what would be the most common response?)}\\[6pt]
A.~\bangla{বলবেন, 'আপনারা তো বিধর্মী, ঈদের দিনে সামনে আসবেন না।'}\\ \textit{(Say, "You are from a different religion, you should not come in front of us on Eid.")}\\
B.~\bangla{হাসিমুখে সেমাই বা খাবারের দাওয়াত দেবেন।}\\ \textit{(Cheerfully invite them for \textit{Semai} or food.)}\\
C.~\bangla{'হ্যাপি হলিডে' বলে নিজের কাজে চলে যাবেন, কারণ এটা তাদের উৎসব নয়।}\\ \textit{(Say, "Happy holiday" and go on with your work, since it is not their festival.)}\\
D.~\bangla{বলবেন, "ঈদ তো আপনাদের উৎসব নয়, তাই টেকনিক্যালি আপনাকে উইশ করাটা লজিক্যাল না।"}\\ \textit{(Say, "Eid is not your festival, so technically it is not logical to wish you.")}
} \\

\midrule

\bottomrule
\end{tabular}
}
\caption{Example instances from the Social Customs Domain. Contd...}
\label{tab:example_prompt_customs_2}
\end{table*}

\clearpage

\begin{table*}[t]
\centering
\large
\resizebox{\textwidth}{!}
{
\large
\begin{tabular}{c|c}
\toprule
\textbf{Subcategory} & \textbf{Example Instances} \\
\midrule

\multicolumn{2}{c}{\textbf{Social Customs}} \\
\midrule
\makecell[l]{
\textbf{Cordiality} \\
}
&
\makecell[l]{
\bangla{সহকর্মীদের সাথে খাচ্ছেন, বিল দেওয়ার সময় হয়েছে।}\\
\textit{(You are eating out with colleagues, and it's time to pay the bill.)}\\[4pt]
\bangla{এই প্রেক্ষাপটে বাংলাদেশী সংস্কৃতিতে সবচেয়ে প্রচলিত উপায়ে আপনি কী করবেন?}\\
\textit{(In this context, according to Bangladeshi culture, which honorific form of address is most appropriate?)}\\[6pt]
A.~\bangla{বলবেন, 'আমরা সবাই মিলে মোট ৩২০০ টাকার খাবার গ্রহণ করেছি।'}\\ \textit{(Say, "All of us together have eaten food worth 3,200 taka.")}\\
B.~\bangla{নিজের খাবারের অংশটুকু হিসাব করে শুধু নিজের টাকাটাই দেবেন।}\\ \textit{(Calculate only your own share of the meal and pay just that amount.)}\\
C.~\bangla{সবার আগে বিলের কাগজ কেড়ে নেবেন এবং 'আমি দিচ্ছি, আমি দিচ্ছি' বলে জোর করবেন}\\\textit{(Grab the bill first and insist, saying, "I'll pay, I'll pay.")}\\
D.~\bangla{সবার বিল দেওয়ার পাশাপাশি ওয়েটারকে বিলের দ্বিগুণ টিপস দিয়ে দেবেন।}\\ \textit{(Pay everyone's bill and also give the waiter a tip equal to twice the bill amount.)}
} \\

\midrule

\makecell[l]{
\textbf{Criticism} \\
}
&
\makecell[l]{
\bangla{কর্মচারী গ্রাহকের সামনে ভুল বিল করেছে।}\\
\textit{(An employee has made a mistake in the bill in front of a customer.)}\\[4pt]
\bangla{এই প্রেক্ষাপটে বাংলাদেশী সংস্কৃতিতে সবচেয়ে প্রচলিত উপায়ে আপনি কী করবেন?}\\
\textit{(In this context, according to Bangladeshi culture, what would be the most common response?)}\\[6pt]
A.~\bangla{গ্রাহকের সামনেই কর্মচারীকে বলবেন, 'গাধা নাকি? যোগ-বিয়োগও পারো না?}\\ \textit{(Scold the employee in front of the customer, saying, "Are you an idiot? Can't you even do basic addition and subtraction?")}\\
B.~\bangla{ভুল বিলটাই নিজের পকেট থেকে দিয়ে দেবেন যাতে কর্মচারীর ভুল ধরা না পড়ে}\\ \textit{(Pay the incorrect bill out of your own pocket so that the employee's mistake is not noticed.)}\\
C.~\bangla{বলবেন, 'বিলের টোটাল অ্যামাউন্টে এরিথমেটিক এরর আছে।'}\\ \textit{(Say, "There is an arithmetic error in the total amount on the bill.")}\\
D.~\bangla{গ্রাহকের কাছে ক্ষমা চেয়ে বিলটি ঠিক করে দেবেন, কর্মচারীকে পরে বোঝাবেন।}\\ \textit{(Apologize to the customer and correct the bill, and explain the mistake to the employee later.)}
} \\

\midrule

\makecell[l]{
\textbf{Time} \\
}
&
\makecell[l]{
\bangla{বিয়ের দাওয়াতে এসে দেখলেন পার্লার থেকে কনে আসার কথা ৮টায়, রাত ১০টায় কনে আসল।}\\
\textit{(The bride was supposed to arrive from the beauty parlor at 8:00 p.m., but she arrived at 10:00 p.m.)}\\[4pt]
\bangla{এই প্রেক্ষাপটে বাংলাদেশী সংস্কৃতিতে সবচেয়ে প্রচলিত উপায়ে আপনি কী করবেন?}\\
\textit{(In this context, according to Bangladeshi culture, what would be the most common response?)}\\[6pt]
A.~\bangla{'বউ সাজতে সময় লাগে' বলে মেনে নেবেন এবং সুন্দর হয়েছে বলে প্রশংসা করবেন}\\ \textit{(Accept it by saying, "Getting dressed as a bride takes time," and compliment her on how beautiful she looks.)}\\
B.~\bangla{বরপক্ষ হলে রেগে গিয়ে বিয়ে ভেঙে দেওয়ার হুমকি দিবেন}\\ \textit{(If you are from the groom's side, get angry and threaten to call off the wedding.)}\\
C.~\bangla{বলবেন, 'মাত্র ২ ঘণ্টা দেরি? ভালো সাজার জন্য তো সারারাত লাগানো উচিত ছিল!'}\\ \textit{(Say sarcastically, "Only a two-hour delay? For proper bridal makeup, it should have taken the whole night!")}\\
D.~\bangla{বলবেন, 'কনের মেকআপের লেয়ার অনেক পুরু।'}\\ \textit{(Say, "The bride's makeup layers are very thick.")}
} \\

\bottomrule
\end{tabular}
}
\caption{Example instances from the Social Customs Domain.}
\label{tab:example_prompt_customs_3}
\end{table*}

\twocolumn

\section{\benchname Development Details}
\label{sec:dev}
\subsection{Annotator Recruitment}
We recruited six annotators who met three strict eligibility criteria: (i) native-level proficiency in Bangla, (ii) lifelong residence in Bangladesh, and (iii) sustained exposure to everyday Bangladeshi social interaction. All annotators reported regular use of Bangla across personal, academic, and professional contexts.

The annotator group reflected diverse educational and experiential backgrounds. Three annotators held undergraduate degrees, two were graduate students with experience in natural language processing, and one had completed a college-level degree. 

To account for religious variation in address forms, kinship terminology, and social customs, the annotator pool included four Muslim annotators and two Hindu annotators.

\subsection{Cultural Situation Seed Collection}

Instance construction began with a cultural situation seeding phase. Annotators were instructed to brainstorm short descriptions of culturally plausible social situations drawn from everyday Bangladeshi life (for example, interactions between parents and children, conversations between superiors and subordinates in office, exchanges among relatives, or encounters between acquaintances and strangers). 

To support cultural authenticity, annotators were explicitly permitted to consult external cultural sources, including Bangladeshi literary works (novels, short stories and dramas) and ethnographic or sociological studies of Bangladeshi society. These materials were used solely to inspire situation design and social dynamics; annotators were instructed not to copy text, character names, dialogue, or narrative structure from any source. All situations were independently written from scratch.

\subsection{Domain-Specific Instance Expansion}

\paragraph{Bangla Address Terms.}
For address term instances, initial situation seeds were expanded by explicitly varying sociopragmatic dimensions such as participant age, gender, key, setting, end \citep{hymesIntroductionEthnographiesCommunication11964}. During this phase, the core taxonomy of Address term domain was defined and refined. Each expanded instance specifies a realistic interactional context in which correct pronominal or nominal address choice depends on these interacting factors rather than on lexical cues.

\paragraph{Kinship Reasoning.}
Kinship instances were developed to test relational reasoning over Bangladeshi kinship systems. Annotators were allowed to consult educational practice materials related to family-tree reasoning solely as conceptual reference. Direct reuse or paraphrasing was prohibited. Each instance was written by constructing a novel relational description and modifying family structures, relations, or generational depth to ensure originality. Correct answers require reasoning over lineage and relational structure rather than memorization of kinship terms.

\paragraph{Social Customs.}
For the Social Customs domain, annotators identified recurrent cultural practices and interactional expectations prevalent in Bangladeshi society, drawing on anthropological studies and cultural reference materials. These practices were first discussed and abstracted into shared cultural scripts \citep{goddardCulturalScriptsWhat2004}, which were then explicated into concrete situational questions. Each instance presents a realistic scenario followed by four response options, only one of which aligns with conventional Bangladeshi expectations regarding appropriate behavior or linguistic response.

\subsection{Instance Validation and Integration}

Following instance writing, all questions were reviewed to ensure clarity, cultural plausibility, and absence of lexical shortcuts. Redundant or weakly grounded instances were removed prior to annotation.

\subsection{Annotator Workload Distribution}

Annotation responsibilities were distributed to balance workload and domain familiarity. Four annotators were assigned two subcategories each, while two annotators were assigned a single subcategory based on stated preference, prior familiarity, and anticipated annotation effort. 

Two experienced annotators additionally served as \emph{lead annotators} and conducted secondary review and adjudication. In cases of disagreement or uncertainty, they evaluated competing annotations, verified cultural plausibility, and finalized labels through discussion. This adjudication process ensured consistency and cultural validity across subcategories.

\section{Annotation Details}
\label{sec:annotation_details}
\subsection{Inter-Annotator Agreement}

To assess the reliability of human annotations in \benchname, we measure inter-annotator agreement (IAA) using Cohen’s $\kappa$ \citep{cohenCoefficientAgreementNominal1960}. Cohen’s $\kappa$ accounts for agreement expected by chance and is widely used for categorical annotation tasks.

\subsection{Standard Cohen’s \texorpdfstring{$\kappa$}{kappa}}

Given two annotators $A_1$ and $A_2$, Cohen’s $\kappa$ is defined as:
\begin{equation}
\kappa = \frac{P_o - P_e}{1 - P_e},
\end{equation}
where $P_o$ denotes the observed agreement between annotators and $P_e$ denotes the expected agreement by chance, estimated from the marginal label distributions.

\subsection{Binary-Decomposed \texorpdfstring{$\kappa$}{kappa} for Multi-Label Annotations}

Pronominal Addressing subset of \benchname allow two valid answers (e.g., \textit{Primary} and \textit{Secondary} address forms). In such cases, direct application of standard Cohen’s $\kappa$ may underestimate agreement, since annotators may select different but equally acceptable labels.

To address this issue, we adopt a \textit{binary decomposition} strategy following prior work on inter-annotator agreement for semantic and dialogue annotation \citep{artsteinSurveyArticleInterCoder2008}.

Let $\mathcal{C} = \{c_1, \dots, c_K\}$ denote the set of possible labels for a given subset. For each label $c_k$, we define a binary indicator variable:
\begin{equation}
y_{i,k} =
\begin{cases}
1 & \text{if annotator $i$ assigns label $c_k$}, \\
0 & \text{otherwise}.
\end{cases}
\end{equation}

Using these binary variables, we compute a separate Cohen’s $\kappa$ for each label $c_k$:
\begin{equation}
\kappa_k = \frac{P_o^{(k)} - P_e^{(k)}}{1 - P_e^{(k)}},
\end{equation}
where $P_o^{(k)}$ and $P_e^{(k)}$ are calculated over the binary label assignments $y_{1,k}$ and $y_{2,k}$.

\subsection{Aggregation Across Labels and Subsets}

For subsets with two acceptable labels, we report a macro-averaged binary $\kappa$:
\begin{equation}
\kappa_{\text{binary}} = \frac{1}{K} \sum_{k=1}^{K} \kappa_k.
\end{equation}

This aggregation treats all labels equally and avoids bias toward high-frequency categories. For subsets with a single-label annotation schema, we report standard Cohen’s $\kappa$. For the full dataset, we additionally report a pooled $\kappa$ computed by concatenating annotator label pairs across all subsets.

The Inter-annotator agreement (Cohens $\kappa$) across annotation categories are reported in Table \ref{tab:iaa_all}.

\clearpage

\begin{table*}[t]
\centering
\normalsize
\renewcommand{\arraystretch}{1.3}
\setlength{\tabcolsep}{14pt}

\begin{tabular}{l l c c}
\hline
\textbf{Category} & \textbf{Subcategory} & \textbf{Standard $\kappa$} & \textbf{Binary $\kappa$} \\
\hline

\multirow{5}{*}{Pronominal Addressing}
& Family             & 0.701 & 0.759 \\
& Office             & 0.670 & 0.768 \\
& Hospital           & 0.792 & 0.872 \\
& Academia           & 0.717 & 0.747 \\
& Social Interaction & 0.637 & 0.721 \\

\hline

\multirow{3}{*}{Nominal Addressing}
& Kinship            & 0.872 & -- \\
& Social Addressing  & 0.808 & -- \\
& Honorifics         & 0.733 & -- \\

\hline

Social Customs & -- & 0.838 & -- \\

\hline
\end{tabular}

\caption{Inter-annotator agreement (Cohen’s $\kappa$) across annotation categories in \benchname. Binary $\kappa$ is reported for pronominal addressing where two valid answers are possible.}
\label{tab:iaa_all}

\end{table*}

\clearpage

\subsection{Experiment Details}
\subsubsection{Zero Shot Evaluation Prompts}
No system-level prompts were used; all task instructions were provided directly within the user prompt
\label{app:zeroshot_prompts}
\paragraph{Prompt Types}

Depending on the annotation schema of each subcategory/subset, we use two prompt formats.

\paragraph{Select any One of Three} For subsets involving second-person pronominal addressing, models were required to output exactly one number from 3 given options. The following prompt shown is translated from actual Bangla Prompt used.

\begin{tcolorbox}[
  colback=brown!5,
  colframe=brown!60!black,
  title=\textbf{Pronominal Addressing Prompt},
  fonttitle=\bfseries
]

\textbf{\bangla{প্রেক্ষাপট}} (\texttt{\small\textbf{Situation}}) : \texttt{<Situation> (e.g., Leaning close to his grandmother, the boy asked why she hadn't told him a story that day.)}  

\textbf{\bangla{প্রশ্ন}} (\texttt{\small\textbf{Question}}): \texttt{<Question> (e.g., In this context, within Bangladeshi social system, how would a grandson typically address his grandmother?)}

\vspace{12pt}
\vspace{2pt}

\textbf{\bangla{বিকল্পসমূহ}} (\small\textbf{Options}):
\begin{enumerate}[label=\textbf{\Alph*.}]
    \item \texttt{\bangla{আপনি} \small(Apni)}
    \item \texttt{\bangla{তুমি} \small(Tumi)}
    \item \texttt{\bangla{তুই} \small(Tui)}
\end{enumerate}

\vspace{6pt}
\bangla{শুধু সঠিক উত্তরের অক্ষরটি লিখুন (A/B/C)।}

\texttt{Respond with the correct option letter only (A/ B/ C)]}
\label{fig:zero_shot_pronominal}
\end{tcolorbox}

\paragraph{Select any One of Four}
For subsets framed as four options, multiple-choice questions, including Nominal Addressing, Social Customs and Kinship Reasoning models are instructed to select exactly one option by outputting the corresponding number.

\begin{tcolorbox}[
  colback=brown!5,
  colframe=darkblue!60!black,
  title=\textbf{Prompt for Multiple-Choice Question},
  fonttitle=\bfseries
]

\textbf{\bangla{প্রেক্ষাপট}} (\texttt{\small\textbf{Situation}}) : \texttt{<Situation> (e.g. While eating, a guest says, "“No, no, please don’t serve any more. I’'m already full."”)}  

\textbf{\bangla{প্রশ্ন}} (\texttt{\small\textbf{Question}}): \texttt{<Question> (e.g. In this context, according to Bangladeshi cultural norms, what would you most commonly say?)} 

\vspace{6pt}
\textbf{\bangla{বিকল্পসমূহ}} (\textbf{Options}):
\begin{enumerate}
    \item \texttt{<Option-1>}
    \item \texttt{<Option-2}
    \item \texttt{<Option-3>}
    \item \texttt{<Option-4>}
\end{enumerate}

\vspace{6pt}
\bangla{শুধু সঠিক উত্তরের নম্বরটি লিখুন (1/2/3/4)।}

\texttt{[Respond with the option number only (1 / 2 / 3 / 4)]}
\label{fig:zero_shot_nominal_et_al}

\end{tcolorbox}

Model outputs are parsed using robust numeric extraction to account for formatting variation. A response is marked correct only if the selected option exactly matches the gold answer.

\subsubsection{Random Baseline Accuracy Computation for Pronominal Addressing}
\label{app:random-baseline}
The pronominal addressing subset of \benchname contains 433 instances specifying a single acceptable pronoun, while 157 instances specifying both a primary and a secondary acceptable pronoun. To account for the differing proportions of these two instance types, we compute an instance-weighted random baseline as follows:

\begin{equation}
\mathrm{Acc}_{\mathrm{random}} =
\frac{433}{590} \cdot \frac{1}{3}
+
\frac{157}{590} \cdot \frac{2}{3}.
\end{equation}

\begin{figure*}
    \centering
    \includegraphics[width=\textwidth]{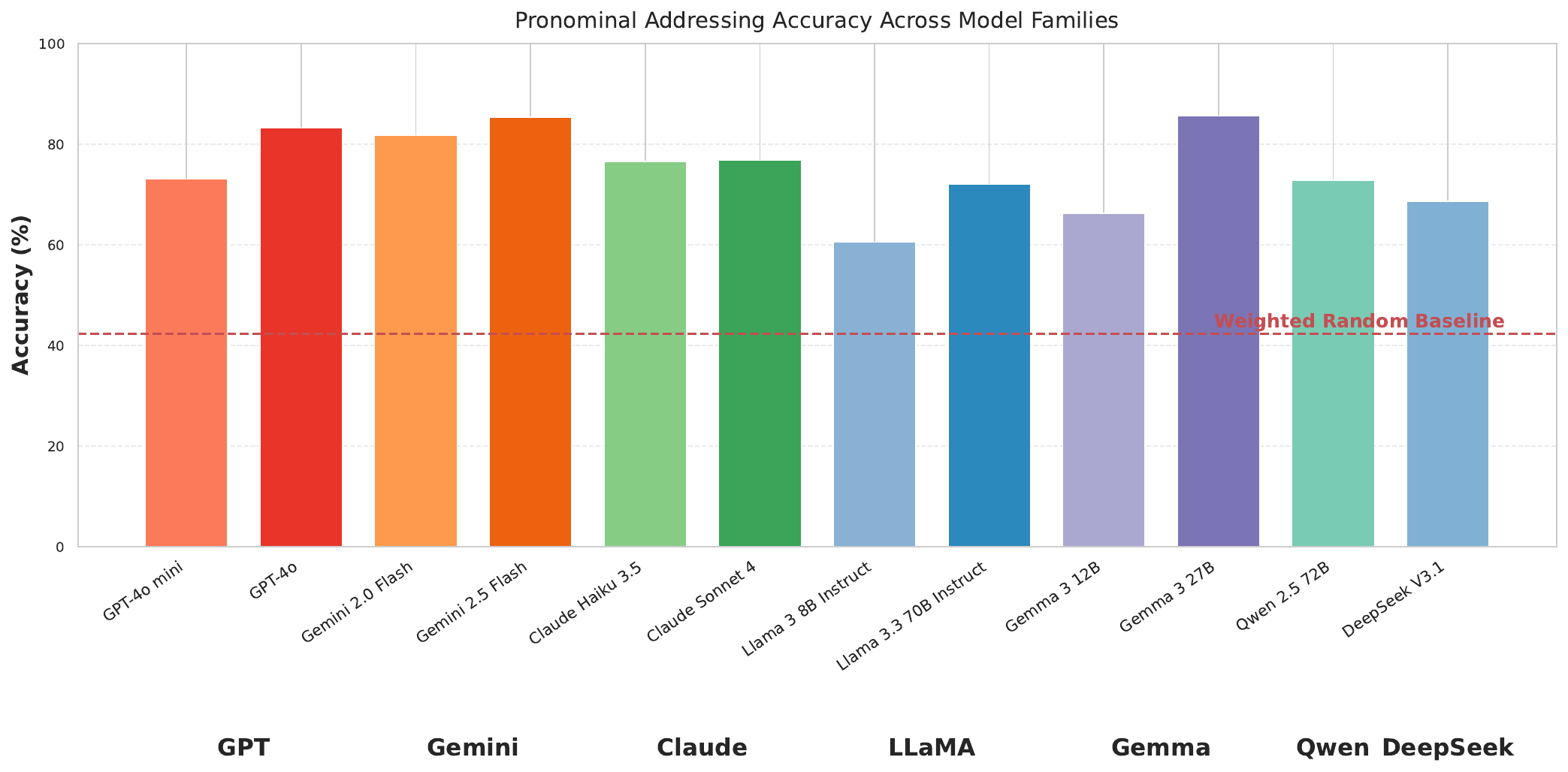}
    \caption{Pronominal addressing accuracy across evaluated language models. Bars indicate model accuracy under the primary-or-secondary evaluation criterion, while the dashed horizontal line denotes the weighted random baseline computed from the distribution of single- and dual-acceptable instances. Models are grouped by family.}
    \label{fig:weighted_random_fig}
\end{figure*}

This yields an expected random accuracy of $42.2\%$ for the pronominal addressing subset. Overall accuracy acrosss Pronominal Addressing is shown in Figure \ref{fig:weighted_random_fig}.

\clearpage

\begin{figure*}
    \centering
    \includegraphics[width=\textwidth]{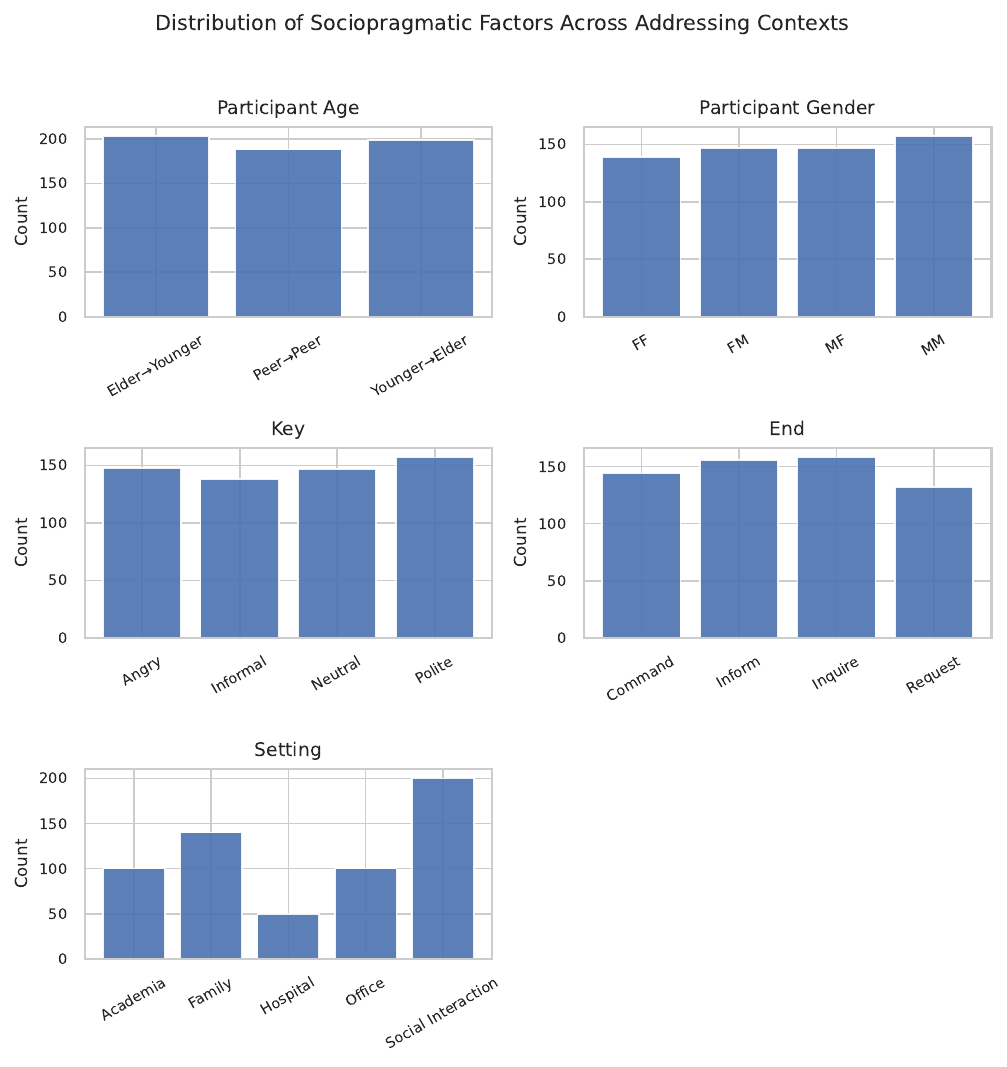}
    \caption{Distribution of sociopragmatic factors used in instance construction across pronominal addressing settings. Each subplot shows the number of instances per category for participant age, gender configuration, politeness key, utterance end type, and interaction setting. The distributions reflect the balanced design used during dataset construction}
    \label{fig:class_balance}
\end{figure*}

\clearpage

\subsubsection{Statistical Analysis of Sociopragmatics Cues and Culturally Inappropiate Pronoun Prediction}

\label{sec:chi_analysis}

To assess whether sociopragmatic factors are systematically associated with model errors, we apply Pearson's $\chi^2$ test of independence between binary error outcomes and categorical sociopragmatic variables.

\subsection{Test Setup}

For each model and each sociopragmatic factor (Participant Age, Participant Gender, Setting, Key, End), we construct a contingency table of the form:
\begin{center}
\begin{tabular}{c|cc}
 & Error & Correct \\
\hline
Category$_1$ & $O_{11}$ & $O_{12}$ \\
Category$_2$ & $O_{21}$ & $O_{22}$ \\
$\vdots$ & $\vdots$ & $\vdots$ \\
\end{tabular}
\end{center}

Let $O_i$ denote the observed count for category $i$, and $E_i$ the expected count under the null hypothesis of independence.

\subsection{Chi-Square Statistic}

The $\chi^2$ statistic is computed as:
\begin{equation}
\chi^2
=
\sum_{i=1}^{K}
\frac{(O_i - E_i)^2}{E_i},
\end{equation}
where $K$ is the number of categories for a given sociopragmatic factor.

Statistical significance is assessed against the $\chi^2$ distribution with $(K-1)$ degrees of freedom. We consider associations significant at $p < 0.05$.

\subsection{Standardized Residuals}

To identify which specific categories drive significant associations, we compute standardized residuals:
\begin{equation}
r_i = \frac{O_i - E_i}{\sqrt{E_i}}.
\end{equation}

Values $|r_i| > 2$ indicate categories that deviate substantially from chance expectations. Positive residuals correspond to higher-than-expected error rates, while negative residuals indicate relative robustness.

\subsection{Sample Size and Validity}

Across all evaluated models, the analysis includes $N = 590$ instances per model. All contingency tables satisfy the expected-frequency condition ($E_i \geq 5$) required for the validity of the $\chi^2$ approximation. No smoothing or reweighting was applied.

\subsection{Findings}

Across all tests, the minimum expected cell count exceeded 5 (min = 14.0), satisfying the assumptions of Pearson’s $\chi^2$ test.

\begin{figure*}
    \centering
    \includegraphics[width=\textwidth]{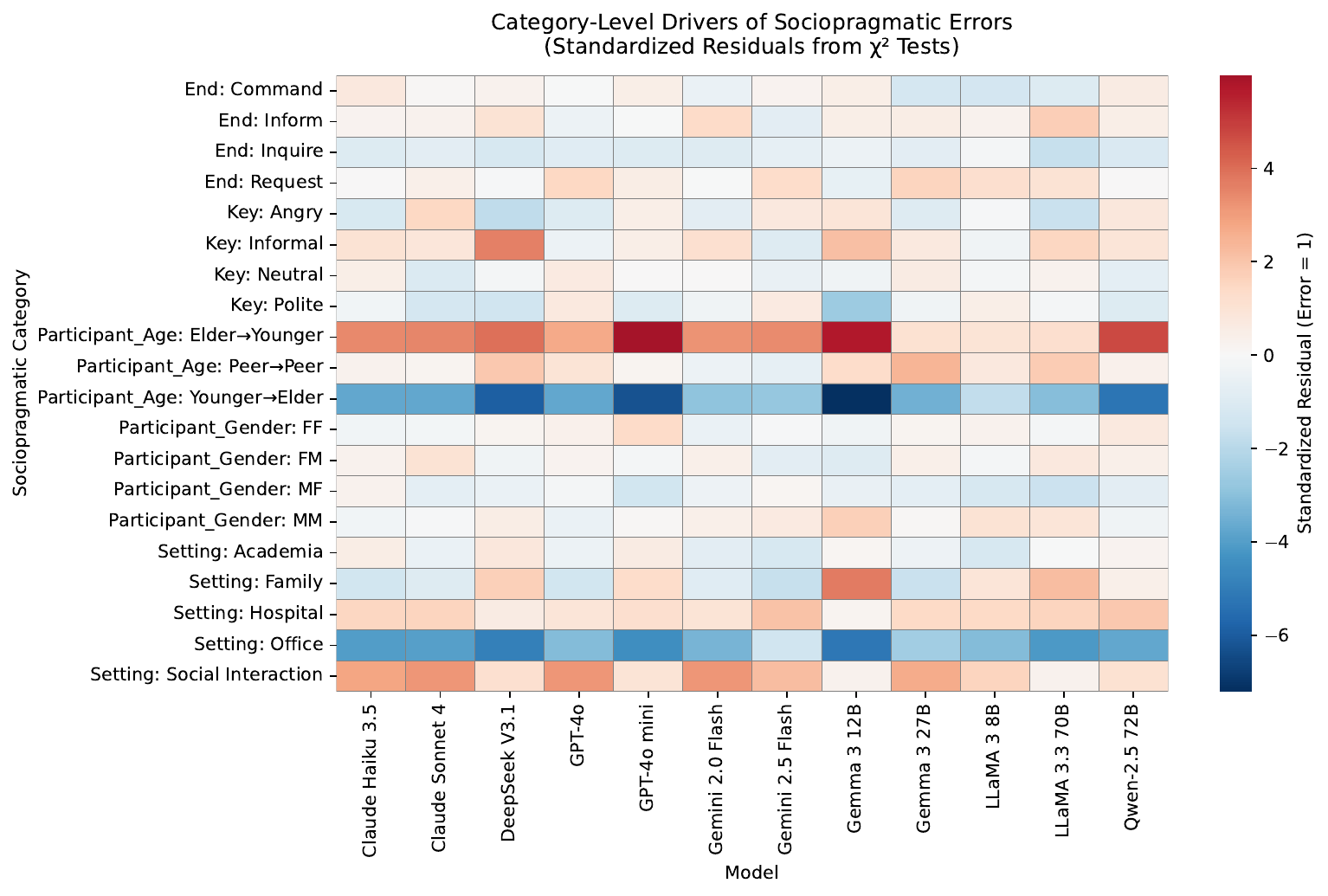}
    \caption{}
    \label{fig:chi_heatmap}
\end{figure*}

As shown in Figure~\ref{fig:chi_heatmap}, participant age direction and interactional setting emerge as the most consistent drivers of sociopragmatic failure across models.
In particular, \textit{Elder$\rightarrow$Younger} interactions and informal social settings are associated with substantially elevated error rates, whereas \textit{Younger$\rightarrow$Elder} interactions and institutional settings (e.g., office, academia) show reduced error incidence.
These patterns align with the main text findings and confirm that sociopragmatic misalignment in LLMs reflects structured sensitivity to social hierarchy and formality rather than uniform noise.

Detailed per-model Pearson's $\chi^2$ test results for all sociopragmatic factors considered are reported in Table~\ref{tab:chi_social}. The table provides model-wise statistics including sample size, chi-square values, degrees of freedom, $p$-values, and minimum expected cell counts, enabling fine-grained inspection of which sociopragmatic dimensions are associated with culturally inappropriate predictions for each model.

\clearpage


\onecolumn
\setlength{\tabcolsep}{4pt}
\begin{longtable}{llrrrrr}
\caption{Per-model Pearson's $\chi^2$ tests of independence between sociopragmatic factors and culturally inappropriate responses.
For each model and predictor, we report the sample size ($N$), chi-square statistic ($\chi^2$), degrees of freedom (DoF),
$p$-value, and the minimum expected cell count. Statistically significant associations ($p < 0.05$) indicate that error
rates are not uniformly distributed across sociopragmatic conditions.}
\label{tab:chi_social} \\

\toprule
\textbf{Model} & \textbf{Predictor} & \textbf{$N$} & \textbf{$\chi^2$} & \textbf{DoF} & \textbf{$p$-value} & \textbf{Min. Exp.} \\
\midrule
\endfirsthead

\multicolumn{7}{c}{\textit{Table \thetable\ continued from previous page}} \\
\toprule
\textbf{Model} & \textbf{Predictor} & \textbf{$N$} & \textbf{$\chi^2$} & \textbf{DoF} & \textbf{$p$-value} & \textbf{Min. Exp.} \\
\midrule
\endhead

\midrule
\multicolumn{7}{r}{\textit{Continued on next page}} \\
\endfoot

\bottomrule
\endlastfoot

LLaMA 3.3 70B    & End                 & 590 & 10.795  & 3   & 0.0128  & 36.92         \\
Gemma 3 27B      & End                 & 590 & 5.913   & 3   & 0.1159     & 19.02         \\
LLaMA 3 8B       & End                 & 590 & 5.53    & 3   & 0.1368    & 52.13         \\
DeepSeek V3.1    & End                 & 590 & 3.756   & 3   & 0.2890     & 41.39         \\
GPT-4o           & End                 & 590 & 3.688   & 3   & 0.2972    & 22.15         \\
Gemini 2.0 Flash & End                 & 590 & 3.661   & 3   & 0.3005     & 24.16         \\
Gemini 2.5 Flash & End                 & 590 & 3.262   & 3   & 0.3530      & 19.24         \\
Qwen-2.5 72B     & End                 & 590 & 2.986   & 3   & 0.3937    & 35.8          \\
GPT-4o mini      & End                 & 590 & 2.204   & 3   & 0.5312     & 35.57         \\
Claude Haiku 3.5 & End                 & 590 & 2.058   & 3   & 0.5604     & 30.87         \\
Claude Sonnet 4  & End                 & 590 & 1.141   & 3   & 0.7673     & 30.65         \\
Gemma 3 12B      & End                 & 590 & 0.814   & 3   & 0.8461     & 44.52         \\
DeepSeek V3.1    & Key                 & 590 & 26.922  & 3   & 6.112e-06  & 43.27         \\
Gemma 3 12B      & Key                 & 590 & 15.257  & 3   & 0.0016   & 46.55         \\
Claude Sonnet 4  & Key                 & 590 & 7.245   & 3   & 0.0644    & 32.04         \\
LLaMA 3.3 70B    & Key                 & 590 & 6.95    & 3   & 0.0735    & 38.59         \\
Claude Haiku 3.5 & Key                 & 590 & 3.34    & 3   & 0.3420      & 32.28         \\
Qwen-2.5 72B     & Key                 & 590 & 2.646   & 3   & 0.4495    & 37.42         \\
GPT-4o           & Key                 & 590 & 2.523   & 3   & 0.4711     & 23.16         \\
Gemini 2.0 Flash & Key                 & 590 & 2.482   & 3   & 0.4786     & 25.26         \\
Gemini 2.5 Flash & Key                 & 590 & 2.441   & 3   & 0.4860    & 20.12         \\
Gemma 3 27B      & Key                 & 590 & 2.17    & 3   & 0.5379     & 19.88         \\
GPT-4o mini      & Key                 & 590 & 1.917   & 3   & 0.5897     & 37.19         \\
LLaMA 3 8B       & Key                 & 590 & 0.653   & 3   & 0.8842     & 54.5          \\
Gemma 3 12B      & Participant\_Age    & 590 & 108.894 & 2   & 2.2591e-24 & 63.41         \\
GPT-4o mini      & Participant\_Age    & 590 & 102.284 & 2   & 6.1562e-23  & 50.66         \\
DeepSeek V3.1    & Participant\_Age    & 590 & 78.349  & 2   & 9.7009e-18  & 58.95         \\
Qwen-2.5 72B     & Participant\_Age    & 590 & 58.254  & 2   & 2.240e-13 & 50.98         \\
Claude Sonnet 4  & Participant\_Age    & 590 & 33.699  & 2   & 4.812e-08 & 43.65         \\
Claude Haiku 3.5 & Participant\_Age    & 590 & 33.629  & 2   & 4.984e-08 & 43.97         \\
GPT-4o           & Participant\_Age    & 590 & 26.561  & 2   & 1.707e-06  & 31.55         \\
Gemini 2.0 Flash & Participant\_Age    & 590 & 23.38   & 2   & 8.377e-06  & 34.41         \\
Gemini 2.5 Flash & Participant\_Age    & 590 & 22.992  & 2   & 1.016e-05  & 27.4          \\
Gemma 3 27B      & Participant\_Age    & 590 & 22.722  & 2   & 1.164e-05  & 27.08         \\
LLaMA 3.3 70B    & Participant\_Age    & 590 & 19.641  & 2   & 5.433e-05  & 52.58         \\
LLaMA 3 8B       & Participant\_Age    & 590 & 7.734   & 2   & 0.0209   & 74.24         \\
LLaMA 3.3 70B    & Participant\_Gender & 590 & 5.46    & 3   & 0.1410     & 38.87         \\
GPT-4o mini      & Participant\_Gender & 590 & 5.259   & 3   & 0.1537    & 37.46         \\
LLaMA 3 8B       & Participant\_Gender & 590 & 4.203   & 3   & 0.2403    & 54.89         \\
Gemma 3 12B      & Participant\_Gender & 590 & 3.47    & 3   & 0.3246    & 46.88         \\
Qwen-2.5 72B     & Participant\_Gender & 590 & 2.041   & 3   & 0.5638     & 37.69         \\
Claude Sonnet 4  & Participant\_Gender & 590 & 2.038   & 3   & 0.5645     & 32.28         \\
Gemini 2.5 Flash & Participant\_Gender & 590 & 1.16    & 3   & 0.7626     & 20.26         \\
DeepSeek V3.1    & Participant\_Gender & 590 & 0.929   & 3   & 0.8185     & 43.58         \\
Gemini 2.0 Flash & Participant\_Gender & 590 & 0.87    & 3   & 0.8327     & 25.44         \\
Gemma 3 27B      & Participant\_Gender & 590 & 0.803   & 3   & 0.8486     & 20.03         \\
GPT-4o           & Participant\_Gender & 590 & 0.504   & 3   & 0.9180     & 23.32         \\
Claude Haiku 3.5 & Participant\_Gender & 590 & 0.396   & 3   & 0.9411     & 32.51         \\
Gemma 3 12B      & Setting             & 590 & 62.986  & 4   & 6.831e-13  & 16.86         \\
DeepSeek V3.1    & Setting             & 590 & 42.365  & 4   & 1.401e-08 & 15.68         \\
Claude Sonnet 4  & Setting             & 590 & 38.507  & 4   & 8.805e-08   & 11.61         \\
Claude Haiku 3.5 & Setting             & 590 & 37.203  & 4   & 1.636e-07 & 11.69         \\
LLaMA 3.3 70B    & Setting             & 590 & 34.451  & 4   & 6.021e-07  & 13.98         \\
GPT-4o mini      & Setting             & 590 & 33.069  & 4   & 1.156e-06 & 13.47         \\
Gemini 2.0 Flash & Setting             & 590 & 29.092  & 4   & 7.488e-06  & 9.15          \\
GPT-4o           & Setting             & 590 & 27.23   & 4   & 1.785e-05 & 8.39          \\
LLaMA 3 8B       & Setting             & 590 & 26.979  & 4   & 2.007e-05 & 19.75         \\
Qwen-2.5 72B     & Setting             & 590 & 26.372  & 4   & 2.662e-05 & 13.56         \\
Gemma 3 27B      & Setting             & 590 & 20.812  & 4   & 0.0003 & 7.2           \\
Gemini 2.5 Flash & Setting             & 590 & 18.203  & 4   & 0.001 & 7.29
\end{longtable}

\twocolumn


\clearpage
\subsubsection{Identification of Cross-Religious Kinship Misalignment}
\label{sec:religious_misalignment}

\paragraph{Discriminative Probing}
To analyze cross-religious misalignment in kinship term prediction, we examine model outputs from both the \textit{Kinship Terms} and \textit{Social Nominal Addressing} subcategories of Bengali Address term subset. 

For the \textbf{explicit} and \textbf{implicit} prompting settings, instances were constructed with a balanced number of prompts indicating either Hindu or Muslim identity. Each instance contains a question followed by two Hindu kinterm options and two Muslim kinterm options, both of which are semantically substitutable \textnormal{(e.g., \bangla{A. পিসি} (\textit{Pishi}), \bangla{B. মাসি} (\textit{Mashi}), \bangla{C. খালা} (\textit{Khala}), \bangla{D. ফুপু} (\textit{Fupu})). Religious context is introduced in two ways:}

\begin{enumerate}[label=(\roman*)]

    \item \textbf{Implicit religious cues}: Religious context is conveyed indirectly through culturally associated personal names (e.g., Hindu or Muslim names), without explicitly stating religious identity.
    
    \item \textbf{Explicit religious cues}: Religious identity is stated directly in the prompt (e.g., explicitly referring to a person as Hindu or Muslim).
\end{enumerate}
All models were instructed to respond from a \textbf{Bangladeshi cultural perspective}.

\paragraph{Generative Probing}
In the generative prompting setting, models were asked to produce an appropriate kinship term or address form without being constrained to predefined options for the same questions as discriminative prompting. Prompts were balanced across three conditions: 

A prediction was labeled as \textit{cross-religiously misaligned} when the model selected or generated a kinship term conventionally associated with a religious group different from that indicated by the prompt context. To identify such cases, we first used GPT-4o as an auxiliary judge to flag potential cross-religious substitutions using curated Hindu and Muslim kinship lexicons. All flagged instances were subsequently manually verified by native Bangla-speaking annotators, and only cases with clear and unambiguous cross-religious substitution were retained for analysis. Disagreements were resolved through discussion.
\begin{tcolorbox}[
    colback=green!8,
    colframe=black!60,
    boxrule=0.6pt,
    arc=2mm,
    left=4pt,
    right=4pt,
    top=6pt,
    bottom=6pt,
    width=\linewidth
]

\textbf{Examples: Implcicit and Explicit Cues under Discriminative Prompting Strategy}

\vspace{6pt}

\textbf{Implicit Muslim context via name}\\
\bangla{বাংলাদেশী সংস্কৃতি অনুযায়ী রাহাত তার বাবার বোনকে প্রচলিত উপায়ে কীভাবে সম্বোধন করবে?} (According to Bangladeshi cultural norms, how would Rahat conventionally address his father’s sister?)\\
\emph{\bangla{a) পিসি (\textit{Pishi}, Father's sister; Hindu)} \\ \bangla{b) মাসি (\textit{Mashi}, Mother's sister; Hindu)} \\ \bangla{c) খালা (\textit{Khala}, Mother's Sister; Muslim}) \\ \bangla{d) ফুপু (\textit{Fupu}, Father's sister; Muslim)}}

\vspace{8pt}

\textbf{Implicit Hindu context via name}\\
বাংলাদেশী সংস্কৃতি অনুযায়ী অর্ণব রায় তার বাবার বোনকে প্রচলিত উপায়ে কীভাবে সম্বোধন করবে? (According to Bangladeshi cultural norms, how would Arnab Roy conventionally address his father’s sister?)\\
\emph{\bangla{a) পিসি (\textit{Pishi}, Father's sister; Hindu)} \\ \bangla{b) মাসি (\textit{Mashi}, Mother's sister; Hindu)} \\ \bangla{c) খালা (\textit{Khala}, Mother's Sister; Muslim}) \\ \bangla{d) ফুপু (\textit{Fupu}, Father's sister; Muslim)}}

\vspace{10pt}

\textbf{Explicit Muslim context}\\
\bangla{বাংলাদেশী সংস্কৃতি অনুযায়ী ইসলাম ধর্মালম্বী কোন ব্যক্তি তার বাবার বোনকে প্রচলিত উপায়ে কীভাবে সম্বোধন করবে? (According to Bangladeshi cultural norms, how would a Muslim person conventionally address his father’s sister?)}\\
\emph{\bangla{a) পিসি (\textit{Pishi}, Father's sister; Hindu)} \\ \bangla{b) মাসি (\textit{Mashi}, Mother's sister; Hindu)} \\ \bangla{c) খালা (\textit{Khala}, Mother's Sister; Muslim}) \\ \bangla{d) ফুপু (\textit{Fupu}, Father's sister; Muslim)}}

\vspace{8pt}

\textbf{Explicit Hindu context}\\
\bangla{বাংলাদেশী সংস্কৃতি অনুযায়ী হিন্দু ধর্মালম্বী কোন ব্যক্তি তার বাবার বোনকে প্রচলিত উপায়ে কীভাবে সম্বোধন করবে}(According to Bangladeshi cultural norms, how would a Hindu person conventionally address his father’s sister?)\\
\emph{\bangla{a) পিসি (\textit{Pishi}, Father's sister; Hindu)} \\ \bangla{b) মাসি (\textit{Mashi}, Mother's sister; Hindu)} \\ \bangla{c) খালা (\textit{Khala}, Mother's Sister; Muslim}) \\ \bangla{d) ফুপু (\textit{Fupu}, Father's sister; Muslim)}}

\end{tcolorbox}

\subsection{Statistical Significance Testing}
\label{sec:religious_stats}

To assess whether observed cross-religious substitution patterns reflect systematic bias rather than random variation, we perform a \textbf{Chi-square goodness-of-fit test} separately for each model. For each model, we aggregate all cross-religious errors across prompting regimes and compare the observed counts of substitutions into Hindu versus Muslim kinship terms against a null hypothesis of uniform distribution.

Formally, let $O = (O_{\text{Hindu}}, O_{\text{Muslim}})$ denote the observed counts of cross-religious substitutions for a given model, and let $E = (E_{\text{Hindu}}, E_{\text{Muslim}})$ denote the expected counts under the null hypothesis, where $E_{\text{Hindu}} = E_{\text{Muslim}} = \frac{1}{2}(O_{\text{Hindu}} + O_{\text{Muslim}})$. The Chi-square statistic is computed as:
\begin{equation}
\chi^2
=
\sum_{i \in \{\text{Hindu}, \text{Muslim}\}}
\frac{\displaystyle (O_i - E_i)^2}{E_i}
\end{equation}

Statistical significance is evaluated using the $\chi^2$ distribution with one degree of freedom. We report exact $p$-values for each model and consider results significant at $p < 0.05$. Models with significant results are interpreted as exhibiting directional cross-religious misalignment in kinship reasoning rather than symmetric or random error behavior. Table~\ref{tab:religious_chi_square_full} reports full chi-square statistics, sample sizes, and significance tests for directional cross-religious kinship misalignment across models.

\begin{table*}[t]
\centering
\small
\setlength{\tabcolsep}{6pt}
\renewcommand{\arraystretch}{1.25}

\begin{tabular}{lrrrrrl}
\hline
\textbf{Model} 
& \textbf{$N$} 
& \textbf{Islam Errors} 
& \textbf{Hindu Errors} 
& $\boldsymbol{\chi^2}$ 
& \textbf{$p$-value} 
& \textbf{Significant} \\
\hline
DeepSeek V3.1              & 56 & 51 & 5  & 37.8 & $7.90\times10^{-10}$ & Yes \\
GPT-4o                     & 47 & 42 & 5  & 29.1 & $6.78\times10^{-8}$  & Yes \\
Claude Sonnet 4            & 49 & 45 & 4  & 34.3 & $4.71\times10^{-9}$  & Yes \\
GPT-4o mini                & 49 & 42 & 7 & 25.0 & $5.73\times10^{-7}$  & Yes \\
LLaMA 3.3 70B Instruct             & 49 & 39 & 10 & 17.2 & $3.43\times10^{-5}$  & Yes \\
LLaMA 3 8B Instruct               & 49 & 38 & 11 & 14.9 & $1.15\times10^{-4}$  & Yes \\
Gemini 2.0 Flash           & 40 & 34 & 6  & 19.6 & $9.55\times10^{-6}$  & Yes \\
Gemma 3 27B                & 48 & 35 & 13 & 10.1 & $1.50\times10^{-3}$  & Yes \\
Gemini 2.5 Flash           & 25 & 21 & 4  & 11.6 & $6.74\times10^{-4}$  & Yes \\
Claude Haiku 3.5           & 65 & 44 & 21 & 8.14 & $4.33\times10^{-3}$  & Yes \\
Qwen 2.5 72B               & 59 & 38 & 21 & 4.90 & $2.69\times10^{-2}$  & Yes \\
Gemma 3 12B                & 24 & 10  & 14 & 0.67 & $4.14\times10^{-1}$  & No  \\
\hline
\end{tabular}

\caption{Directional asymmetry in cross-religious kinship term substitutions.
This table reports the total number of culturally inappropriate Kinship Term predictions ($N$) where a model substituted a kinship term from one religious context with another (e.g., using a Hindu-specific term in an Islamic context). We define \textit{Islam Errors} as instances where a model predicts a Hindu kinterm in a given Muslim context, and \textit{Hindu Errors} as the vice versa. Results are aggregated across explicit, implicit, and generative prompting regimes. Significance is determined by a $\chi^2$ goodness-of-fit test ($df=1$), testing the null hypothesis of an equal distribution ($50/50$ split) between Islam and Hindu error types. Results with $p < 0.05$ indicate a statistically significant directional substitution.}
\label{tab:religious_chi_square_full}
\end{table*}

\clearpage

\subsubsection{Log-Probabilities of Human Preferred Primary and Secondary Address Pronouns}
\label{logprob}

For a subset of evaluated models, token-level log probabilities over candidate responses are available. We leverage these log probabilities to analyze how models distribute probability mass across multiple sociopragmatically acceptable responses, specifically the human-annotated \emph{primary} and \emph{secondary} address forms in pronominal addressing.

Log probabilities were obtained from two sources. For proprietary models, we retrieved token-level log probabilities for GPT-4o and GPT-4o-mini using the OpenAI API. For open-weight models, log probabilities were collected through the TogetherAI inference API, including LLaMA~3~8B Instruct, LLaMA~3.3~70B Instruct, Qwen~2.5~72B Instruct, and DeepSeek~V3.1. For each generation step, we retrieve the top-$k$ candidate tokens with $k=5$, corresponding to the five most probable tokens for the first generated output token.

To minimize tokenization inconsistencies across models, answer options were represented using English letter labels (A, B, C), and LLMs were instructed to output only the corresponding letter. This ensured consistent token boundaries when computing probabilities using choice-restricted softmax.

\subsection{Choice-Restricted Softmax}

To obtain comparable probabilities over the candidate pronouns, we apply a restricted softmax normalization over the available choices in $\mathcal{C}$. For each choice $c$, we compute

\begin{equation}
P(c) =
\begin{cases}
\dfrac{\exp(\ell_c)}
{\sum\limits_{c' \in \mathcal{C},\, \ell_{c'}\ \text{defined}} \exp(\ell_{c'})}
& \text{if } \ell_c \text{ is available}, \\
0 & \text{otherwise}.
\end{cases}
\end{equation}

This normalization redistributes probability mass only among the candidate pronouns explicitly considered by the task, without incorporating probability assigned to tokens outside the choice set.

\subsection{Primary and Secondary Probabilities}

Each instance is annotated with a human-preferred \emph{primary} form $c_p$ and, when applicable, a \emph{secondary} acceptable form $c_s$. Using the normalized probabilities defined above, we compute

\begin{equation}
P_{\text{Primary}} = P(c_p), \qquad
P_{\text{Secondary}} = P(c_s).
\end{equation}

To examine how probability mass is distributed across the candidate pronouns, we further compute the maximum probability assigned to any option:

\begin{equation}
P_{\max} = \max_{c \in \mathcal{C}} P(c).
\end{equation}

This value represents the largest share of probability mass allocated to a single candidate pronoun. In later analysis, we compare the distribution of $P_{\max}$ between unambiguous instances (single acceptable answer) and sociopragmatically ambiguous instances (two acceptable answers) to examine whether models distribute probability mass differently when multiple culturally acceptable responses exist.

\end{document}